%% file: 0_main.tex
\documentclass[11pt]{article}

\usepackage[preprint]{acl}

\usepackage{times}
\usepackage{latexsym}

\usepackage[T1]{fontenc}

\usepackage[utf8]{inputenc}

\usepackage{microtype}

\usepackage{inconsolata}

\usepackage{graphicx}

\usepackage{comment}
\usepackage{amsmath}
\usepackage{xcolor}
\usepackage{tcolorbox}
\usepackage{soul} 
\usepackage{adjustbox}
\usepackage{listings}
\usepackage{booktabs} 
\usepackage{amssymb}
\usepackage{tabularx} 
\usepackage{graphicx}
\usepackage{colortbl}
\usepackage{color,soul}
\usepackage[normalem]{ulem}
\usepackage{placeins}
\usepackage[table]{xcolor}
\usepackage{array}
\usepackage{subcaption}

\usepackage{tikz}
\usetikzlibrary{arrows.meta,shapes.arrows,positioning,calc}

\usepackage{wrapfig}
\usepackage{lipsum} 
\usepackage{multicol}
\usepackage{aliascnt}
\usepackage{adjustbox}
\usepackage{booktabs}
\usepackage{multirow} 
\usepackage{hhline}     
\usepackage{subcaption}
\usepackage[inline]{enumitem}
\usepackage{tcolorbox}
\usepackage{arydshln}

\newcolumntype{Y}{>{\color{cGray}\footnotesize\setlength{\parskip}{2pt}\setlength{\parindent}{0pt}}X}
 
\renewcommand{\arraystretch}{1.0}

\tcbuselibrary{skins} 
\newtcolorbox{dialogbox}{
    enhanced,
    boxrule=0.8pt,
    colback=black!5,
    colframe=black,
    left=2pt,
    right=2pt,
    top=2pt,
    bottom=2pt,
    boxsep=1pt,
    fontupper=\ttfamily\scriptsize 
}

\title{HealthLoopQA: A Context-Aware Question Answering Benchmark for Interpreting Wearable Monitoring Data in Diabetes Care}

\author{
 \textbf{Yuchen Niu\textsuperscript{1 *}},
 \textbf{Yanan Ma\textsuperscript{2, 4 *}},
 \textbf{Srinivasan Nandakumar\textsuperscript{5}},
 \textbf{Maolin Chen\textsuperscript{3, 5}},
\\
 \textbf{Viktor Schlegel\textsuperscript{4, 5}},
 \textbf{Kexin Wei\textsuperscript{5}},
 \textbf{Ling Cheng\textsuperscript{5}},
 \textbf{Anna Bird \textsuperscript{5}},
\\
 \textbf{Anil Anthony Bharath\textsuperscript{5}},
 \textbf{Siew-Kei Lam\textsuperscript{1}}
\\
\\
 \textsuperscript{1}Nanyang Technological University,
 \textsuperscript{2}Mayo Clinic,
 \textsuperscript{3}National University of Singapore
  \\
 \textsuperscript{4}The University of Manchester,
\textsuperscript{5}Imperial College London, Imperial Global Singapore
\\
 \textsuperscript{*}These authors contributed equally to this work.
\\
 \small{
   \textbf{Correspondence:} \href{mailto:email@domain}{yuchen017@e.ntu.edu.sg}
 }
}

\definecolor{stagebg}{HTML}{E8EEF7}   
\definecolor{stagefg}{HTML}{1F4E79}   
\definecolor{lblfg}{HTML}{2E7D32}     
\definecolor{codefg}{HTML}{333333}    
\definecolor{cGreen}{HTML}{2E7D32}
\definecolor{cBlue}{HTML}{1F4E79}
\definecolor{cBlueBg}{HTML}{E8EEF7}
\definecolor{cGray}{HTML}{333333}
\definecolor{cOrange}{HTML}{D2691E}

\newcommand{\sect}[1]{{\footnotesize\bfseries\textcolor{lblfg}{#1}}\hspace{0.6em}}
\newcommand{\code}[1]{{\ttfamily\fontsize{9}{8.2}\selectfont\textcolor{codefg}{\def\_{\textunderscore\allowbreak}#1}}}
\newcommand{\stage}[1]{\rowcolor{stagebg}\multicolumn{2}{|c|}{%
  \rule{0pt}{2.4ex}\footnotesize\bfseries\textcolor{stagefg}{#1}}}
\newcolumntype{C}[1]{>{\raggedright\arraybackslash}p{#1}}

\begin{document}
\maketitle
\begin{abstract}
As medical wearables become integrated into daily chronic disease care, effectively interpreting longitudinal monitoring data is essential for patients and clinicians to understand health trends, detect safety-critical events, and make informed decisions. While large language models (LLMs) show promise for transforming this streaming physiological data into personalized health insights, evaluating their reasoning capability and analytical rigor in diverse monitoring tasks remains a fundamental challenge. Existing medical wearable question answering (QA) benchmarks primarily assess short-horizon classification or statistical summaries, largely ignoring the long-term patterns, therapeutic and behavioural contexts, and potential system failures inherent in real-world deployments. To address this, we introduce \textbf{HealthLoopQA}, a comprehensive diagnostic benchmark for evaluating LLM reasoning over continuous diabetes monitoring data. Grounded in a novel taxonomy of eleven atomic reasoning abilities, HealthLoopQA comprises 127 tasks and over 1,500 QA instances spanning process mining, anomaly detection, and prediction over 30-day horizons. To systematically evaluate safety awareness, we complement real-world datasets with a fault-injected simulation testbed modeling diverse device malfunctions and cyber-physical attacks to generate physiologically plausible hazard scenarios. Evaluating state-of-the-art LLMs across prompting and agentic frameworks reveals severe limitations in complex temporal pattern mining. Furthermore, we identify a broader phenomenon of \textit{In-context Laziness} under long-context prompting, highlighting critical open challenges in deploying LLMs for rigorous long-horizon medical reasoning.
\footnote{The code and dataset are publicly available at:
\url{https://github.com/schlevik/py-mgipsim/tree/merge}}

\end{abstract}

\input{1_introduction}

\input{2_related_work}

\input{3_organization_of_benchmark}

\input{4_data_collection}
\input{5_benchmark}
\input{7_conclusion}

\bibliography{main}

\appendix
\input{9_appendix}

\end{document}

%% file: 1_introduction.tex
\section{Introduction}

Out of millions of people with diabetes, many are increasingly relying on continuous glucose monitors (CGMs) and automatic insulin delivery (AID) systems, i.e., wearable physiological closed-loop control systems that automatically adjust insulin delivery based on continuous blood glucose (BG) measurements to maintain BG in a target range, to reduce self-management burden and improve clinical outcomes~\citep{collyns2021improved, Renard2022Automated, Godoi2023Glucose}. The monitoring data collected from these systems are essential for assessing therapy effectiveness, identifying hazards, and informing proactive adjustments~\citep{millson2020analyse, kapadia2024failure}. Yet, the high-volume data stream (e.g., BG measures every 5 minutes) and complex interactions among physiology, therapy, and behavioral context pose challenges for patients and clinicians in meaningfully and effectively interpreting~\citep {mackett2023patient}.

Traditional physiological signal analysis tasks, like classification and forecasting, target narrow objectives and fail to provide open-ended insights.  Recent work has explored Large Language Models (LLMs)-based medical wearable monitoring assistants (MWMAs) through prompting, fine-tuning, and agentic frameworks, demonstrating encouraging performance in translating physiological signals into personalized health insights~\citep{chan2024medtsllm, Liu2023Large, kim2024health, bohi2024large, cosentino2024towards, pawana2025lightweight, khasentino2025personal, feli2025An, merrill2024transforming, fang2024physiollm}.

Despite this potential, there is a critical gap in benchmark tools to evaluate their reasoning capabilities and analytical rigor in diverse monitoring tasks, which are critical to safely guide clinical decisions. Existing medical wearable question-answering (QA) benchmarks primarily assess statistical summaries or short-horizon classification tasks, requiring limited reasoning over the long-term temporal dynamics ~\citep{ecg-qa, llm-cgm}. They also largely ignore the contextual information (e.g., therapy and patient activity) that is essential for interpreting physiological measurements. Furthermore, they rarely account for system failures or adversarial conditions that may arise in real-world deployments and lead to potentially life-threatening errors, as highlighted by prior work and regulatory bodies such as the U.S. FDA and European Medicines Agency ~\citep{kapadia2024failure, niu2025securing,ludvigsen2023role}. 

\begin{figure*}[!t]
    \centering
    \includegraphics[width=1.0\linewidth]{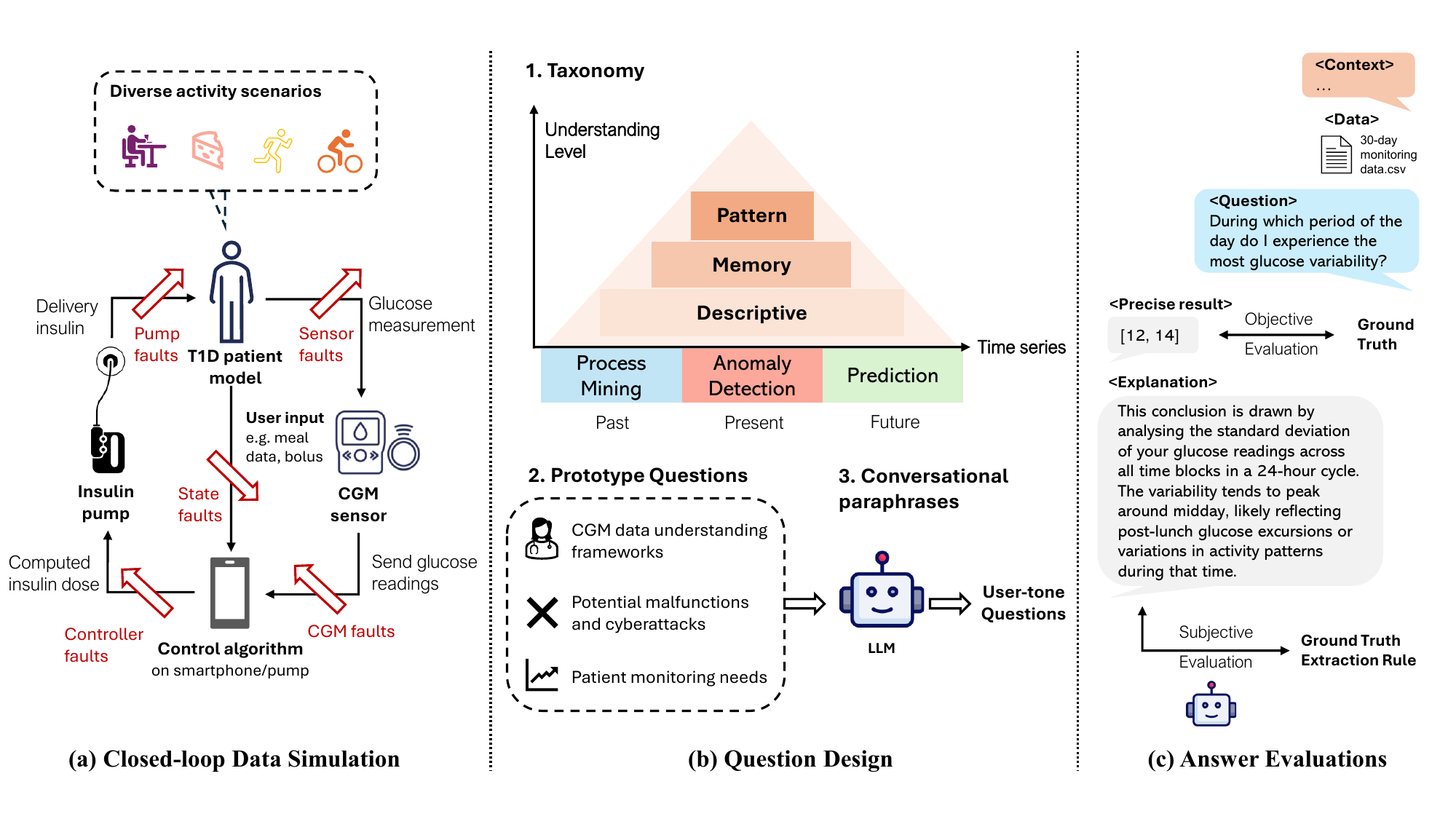}
    \caption{Overview of the HealthLoopQA benchmark. (a) Closed-loop data simulation: simulates the closed loop between virtual T1D patients, CGM sensors, control algorithms, and insulin pumps, with fault modelling to reproduce possible malfunctions and interferences under diverse activity scenarios (e.g., meals, snacks, running, cycling). (b) Question design: a two-dimensional taxonomy is categorised by time-series task type and reasoning depth. Prototype questions are derived from CGM analysis frameworks, clinical guidelines, documented AID vulnerabilities, and patient monitoring needs. Then they are transformed into user-tone questions with LLMs. (c) Answer evaluation:  reasoning instructions and automatic answer-extraction modules paired with each question support evaluation based on numerical precision and reasoning alignment.}
    \label{fig:overview}
\end{figure*}

To address these gaps, this paper introduces HealthLoopQA, an extensible diagnostic benchmark for evaluating reasoning over long-horizon diabetes monitoring data with context and fault injection (Figure~\ref{fig:overview}). In AID systems, precise insulin delivery requires contextual inputs such as manual bolus injection, meal intake, and physical activity. This provides rich real-world long-term physiological monitoring data with therapy and behavioral records \citep{wolff2026metabonet}. Besides, the availability of open-source closed-loop AID testbeds, incorporating clinically validated T1D patient simulators and control algorithms, enables the safe simulation of rare and high-risk system failures with physiologically plausible trajectories representative of the T1D population under diverse meal and physical activity settings \cite{siket2025py}. Together, this work makes four main contributions:

\begin{itemize}
    \item	\textbf{Taxonomy}. We propose a task taxonomy that organizes MWMA evaluation into three analysis stages and eleven atomic reasoning abilities required for interpreting longitudinal wearable data.
    \item \textbf{Fault-injected simulation testbed}. We develop an open-source fault-injected AID simulation testbed based on \cite{siket2025py}, modelling eight device malfunctions and ten cyber-physical attack patterns to enable systematic evaluation of safety and security awareness.
    \item \textbf{HealthLoopQA benchmark}. We curate 127 diagnostic tasks spanning process mining, anomaly detection, and prediction over 30-day monitoring horizons, each accompanied by programmatic ground-truth extraction modules, reasoning instructions, and fine-grained taxonomy annotations. The benchmark contains 1,516 QA instances derived from both simulated and real-world datasets.
    \item \textbf{Comprehensive evaluation}. We benchmark state-of-the-art LLMs using both prompting-based and agentic inference frameworks, revealing critical limitations in longitudinal physiological reasoning, including failures in complex temporal pattern mining and an \emph{``in-context laziness''} phenomenon under long-context prompting.
\end{itemize}

%% file: 2_related_work.tex
\section{Related Work}

\textbf{QA Benchmarking in Time Series Interpretation.} 
QA has been extensively studied in language and vision, but its application to time-series interpretation is relatively recent, enabled by advances in LLMs. Early work such as DeepSQA \citep{xing2021deepsqa} relied on template-based questions with limited linguistic diversity and reasoning difficulty. TimeSeriesExam \citep{cai2024timeseriesexam} introduced 5 core tasks with increasing difficulty to assess LLMs’ general time-series reasoning capabilities, yet it does not involve domain-specific tasks. SensorQA \citep{reichman2025sensorqa} designed human-centred queries over long-duration data in daily-life monitoring, but mainly focused on process mining and evaluated responses using language-similarity metrics rather than numerical accuracy. MTBench \citep{chen2025mtbench} introduced regression and classification metrics for financial and weather data, yet omitted failure-aware monitoring required in medical wearables. 

In healthcare, Time-MQA \citep{kong2025time} covered multiple domains including physiological data, but posed generic questions not tailored to medical wearables monitoring. Domain-specific efforts such as ECG-QA \citep{ecg-qa} and LLM-CGM \citep{healey2024llm}, designed QA for electrocardiogram (ECG) and CGM monitoring respectively, remain limited to physiological signal analysis, lack therapeutic or activity context, and do not systematically model system failures. 

\textbf{Understanding and Analysis of CGM Data.} 
Clinical guidelines such as the 2025 American Diabetes Association (ADA) Standards of Care emphasize CGM-derived metrics, including the Ambulatory Glucose Profile (AGP) and Time in Range (TIR) as key measures for diabetes management \citep{american2025summary}. Statistical and machine learning approaches, including glucodensity curves and long short-term memory (LSTM) models, capture temporal trends and support clustering or forecasting, but often lack interpretability \citep{klonoff2025cgm}. More advanced architectures such as AttenGluco \citep{farahmand2025attengluco} integrate environmental data through cross-attention for long-term forecasting, while GlucoBench \citep{sergazinov2024glucobench} introduces CGM-specific standards for prediction and uncertainty estimation. However, they remain limited in satisfying diverse monitoring needs.   
Recent reviews, such as CGM Data Analysis 2.0 \citep{klonoff2025cgm}, argue that traditional statistical analyses oversimplify dynamic glucose fluctuations and highlight alternative frameworks, including functional data analysis, AI/ML, and foundation models, to enable richer interpretation of complex glucose patterns and support personalized decision-making. This underscores the need for QA tasks that move beyond traditional metrics to capture dynamic glucose trajectories and individualized treatment contexts.


\textbf{QA as Task vs. QA as Diagnostic Tool}.
To establish LLMs' abilities to reason over CGM data, we rely on QA as a diagnostic tool~\citep{srivastava2023beyond}, rather than performing costly human-centred experiments or evaluating task performance by offline proxy tasks~\citep{bedi2025medhelm}, that are mired with common NLG evaluation pitfalls~\citep{Gatt2018,huang2021factual} and not always predict application performance~\citep{Doshi-Velez2017TowardsLearning}. 
Specifically, we focus on the fundamental reasoning abilities that govern understanding, analysis, and inference, 
including physiological reasoning about glucose dynamics, temporal reasoning for pattern recognition in time-series data, and contextual integration of patient-specific factors. Natural language queries serve as the natural interface through which we probe these reasoning capabilities in LLMs. Thus we rely on QA as a task \emph{format}~\citep{gardner2019question} rather than the task itself, where linguistic diversity and clinical plausibility of the queries would be more central.


%% file: 3_organization_of_benchmark.tex
\section{HealthLoopQA}

\begin{figure*}[!t]
    \centering
    \makebox[0.95\textwidth][r]{%
    \resizebox{\dimexpr0.95\textwidth+14pt\relax}{!}{\input{figures/framework}}}
    \caption{The Medical Wearables Monitoring Assistant (MWMA) evaluation framework consists of three stages with different requirements: (1) Understand Intention: interpret user queries and instructions; (2) Reason: break down the instruction in a combination of atomic reasoning steps situated at three distinct cognitive levels;
    (3) Act on task:  execute steps to arrive at final answer.
    }
    \label{fig:task}
\end{figure*}
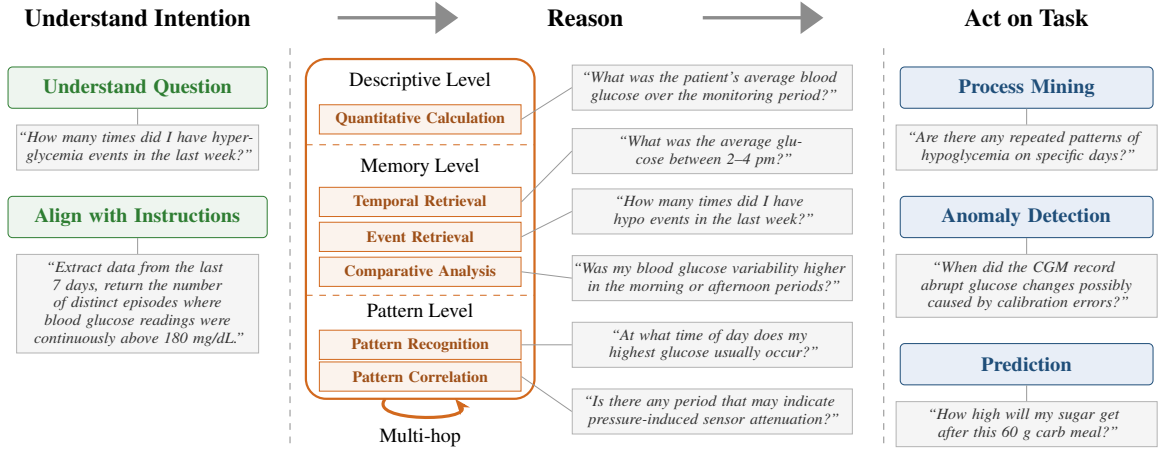

The task is a single-turn QA interaction between a user and an AI assistant. The user provides the following inputs: $X$, the history of monitored physiological time-series data (e.g., 30 days of CGM readings); $C$, therapy and contextual information (e.g., insulin delivery records, meals, and activity logs); $Q$, a natural language question regarding the monitoring data; $I$, a further detailed instruction describing the exact nature of the expected answer to avoid potential ambiguities. 
The AI assistant system $f$ is expected to output $A$, a precise answer, which can take one of several forms depending on the query: \emph{(i)} a precise numerical value, \emph{(ii)} a categorical class label, or \emph{(iii)} a temporal attribute such as a timestamp or event duration, and $R$, the reasoning process in natural language.
We therefore formalise the task as: $f: (X, C, Q, I) \rightarrow (A, R)$.

\subsection{Taxonomy}
We propose a task framework that assesses the ability of MWMAs to analyse CGM with therapy and context data, as shown in Figure~\ref{fig:task}. It contains three analytical stages: Understand Intentions, Reason, and Act on Task. 

To understand the intention,  the evaluated system needs to correctly infer the input objective, task type, and constraints from the query, which is necessary for downstream performance and can be measured by the alignment between the model's reasoning process and the provided instruction.

Reasoning diagnosis is divided into three cognitive levels, inspired by the ``levels of processing'' theory \cite{craik1972levels}: 
\begin{enumerate*}[label=(\emph{\roman*})]
    \item \textbf{Descriptive}: evaluates whether the model can accurately describe physiological time-series data and perform basic numerical or statistical analyses (e.g., average BG).
    \item \textbf{Memory}: evaluates the model’s ability to retrieve relevant monitoring data conditioned on temporal or event-based cues in the query (e.g., ``last Sunday night'' or ``after breakfast'').
    \item \textbf{Pattern}: evaluates whether the model can identify recurrent or latent patterns, detect anomalies, or recognise clinically meaningful patterns described in the query (e.g., nocturnal hypoglycemia).
\end{enumerate*} Under these cognitive levels, reasoning is decomposed into several atomic capabilities including \emph{Quantitative Calculation} (QC), \emph{Temporal Retrieval} (TR), \emph{Event Retrieval} (ER), \emph{Comparative Analysis} (CA), \emph{Pattern Recognition} (PR), and \emph{Pattern Correlation} (PC). They can be composed to support increasingly complex and multi-hop analytical tasks in benchmark questions.  Detailed definitions are in Appendix~\ref{app:atomic}, Table~\ref{tab:atomic_reasoning_supp}. 

Based on collected evidence, the final analytical stage is to act on given tasks to derive answers. We consider core time-series reasoning tasks including 
\begin{enumerate*}[label=(\emph{\roman*})]
    \item \textbf{Process Mining} (PM): retrospectively analyzes monitoring data to identify BG trends, assess therapy effectiveness, and detect patterns such as exercise or postprandial glucose variation.
    \item \textbf{Anomaly Detection} (AD): identifies hazards, manipulated BG readings, and dangerous treatment errors caused by device failures or cyber-physical threats.
    \item \textbf{Prediction} (PD): models future states (e.g., BG levels or insulin needs) supports proactive adjustments in insulin dosing, meal planning, or physical activity to prevent hypo- or hyperglycemia.
\end{enumerate*}

\subsection{Tasks}
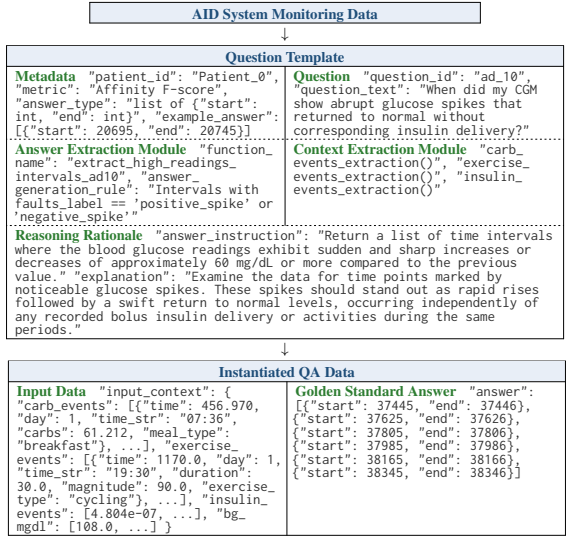
\begin{figure}[!t]
    \centering
    \resizebox{\columnwidth}{!}{\input{figures/data-overview}}
    \caption{Overview of the QA data generation pipeline. Given raw formatted AID monitoring data,  QA instances are automatically produced using the answer and context extraction modules in question templates. }
    \label{fig:qa_template}
\end{figure}

HealthLoopQA comprises 127 distinct tasks that cover the proposed task taxonomy, carefully curated and cross-verified in collaboration with experts in time series analysis, cybersecurity, NLP and diabetes care. The queries and reasoning rationales are designed based on established CGM data analysis practices~\citep{millson2020analyse}, clinical diabetes management guidelines~\citep{american2025summary}, and safety and security monitoring requirements for AID systems~\citep{kapadia2024failure,niu2025securing}. 
The reasoning chain of each question is manually analyzed and labelled based on atomic reasoning capability.
All task instances are generated from templates (Figure \ref{fig:qa_template}) to allow scalable and verifiable generation given AID monitoring data. 


\subsection{QA Data Collection}

\input{tables/qa_data_distribution}

Templates can be instantiated with both real and simulated CGM data. We treat simulated and real-world datasets as complementary evaluation settings: simulation enables controlled assessment of diverse reasoning and safety-critical failure scenarios (e.g., prolonged over-dosage of insulin would lead to hypoglycaemia) that otherwise would be unethical to conduct on human participants, whereas real-world data evaluate robustness under naturally occurring physiological and behavioural variability.

For simulated data, we generate 30-day monitoring trajectories for 16 virtual patients (T1DM Extended Hovorka)~\citep{rashid2019simulation} using the OpenAPS controller for basal insulin delivery with preprandial bolus injection. Daily routines include three meals, two snacks, and one to two exercise events, with timing and magnitude uniformly sampled from plausible ranges. To support anomaly detection tasks, 17 fault patterns are injected into the closed-loop simulation. The simulated dataset supports all 127 benchmark tasks. 

For the real-world dataset, we select 10 months of monitoring data from five individuals with T1D, aged 3--59 years, from the MetaboNet dataset~\citep{wolff2026metabonet}, including recorded carbohydrate intake and physical activity. A subset of 44 suitable tasks that do not require counterfactual data alteration was selected from the process mining and prediction tasks for evaluation on the real-world dataset. While counterfactual data alteration can be readily performed in simulation, it is difficult to apply reliably to real-world data because the true outcomes following an altered state are unobserved and cannot be directly verified. 

Finally, 1,044 QA pairs are randomly sampled from the 16 virtual patients to maintain an approximately balanced distribution across time series tasks, together with 472 samples from the real-world dataset, contributing to an overall 1,516 QA instances for evaluation. 
The task distribution is summarized in Table~\ref{tab:qa_dataset_distribution}. Additional data details regarding question types, simulation testbed and settings, fault modelling, glucose statistics, and QA data schemas are provided in Appendix~\ref{app:data_details}.

\subsection{Quantitative Evaluation Metrics}
\textbf{Regression.} For questions requiring the prediction of continuous numerical values, such as glucose levels or insulin delivery, we measured accuracy using symmetric mean absolute percentage error (SMAPE), defined as 
$\mathrm{SMAPE} = \frac{1}{n}\sum_{i=1}^{n}\frac{\lvert y_i - \hat{y}_i \rvert}{(\lvert y_i \rvert + \lvert \hat{y}_i \rvert)/2}$, 
where $\hat{y}_i$ denotes the model prediction and ${y}_i$ denotes the ground truth. SMAPE normalizes errors relative to the scale of the values, making it suitable for aggregating performance across variables with different ranges.  

\noindent\textbf{Category Classification.} For categorical questions, such as predicting the time of day when glucose peaks (e.g., \textit{morning}, \textit{afternoon}, \textit{evening}, \textit{night}), we reported classification accuracy (ACC).

\noindent\textbf{Event Detection.} For event-related questions with timestamp answers, such as detecting hypoglycemia episodes or abnormal sensor patterns, we used the Affinity F1-score \cite{huet2022local} (F1), which assesses the temporal overlap and alignment between predicted and true event ranges.

ACC and F1 are used directly, while SMAPE, a lower-is-better metric with range $[0,2]$, is scaled as $N_{\mathrm{SMAPE}} = 1 - \frac{\mathrm{SMAPE}}{2}$. With all metrics into $[0,1]$, where larger values indicate better performance, the overall performance score is averaged across questions.

%% file: figures/framework.tex
\definecolor{cGreen}{HTML}{2E7D32}
\definecolor{cBlue}{HTML}{1F4E79}
\definecolor{cBlueBg}{HTML}{E8EEF7}
\definecolor{cGray}{HTML}{333333}
\definecolor{cOrange}{HTML}{D2691E}

\begin{tikzpicture}[
  font=\small,
  greenbox/.style={rectangle, rounded corners=2pt, draw=cGreen, fill=cGreen!8,
                   align=center, inner sep=4pt, minimum height=6.5mm,
                   font=\footnotesize\bfseries, text=cGreen},
  bluebox/.style={rectangle, rounded corners=2pt, draw=cBlue, fill=cBlueBg,
                  align=center, inner sep=4pt, minimum height=6.5mm,
                  font=\footnotesize\bfseries, text=cBlue},
  opbox/.style={rectangle, draw=cOrange, fill=cOrange!8, align=center,
                font=\scriptsize\bfseries, text=cOrange!75!black,
                inner sep=2.5pt, minimum height=4.6mm},
  qbox/.style={rectangle, draw=black!30, fill=black!4, align=center,
               font=\scriptsize\itshape, text=cGray, inner sep=3pt},
  hdr/.style={font=\bfseries},
  lvl/.style={font=\small},
  flowarrow/.style={line width=1.5pt, black!45,
                    -{Triangle[length=3.6mm,width=5mm]}},
  conn/.style={draw=black!40, line width=0.4pt},
]

\def\dA{4.35}  \def\dB{13.7}
\def\sepT{6.7}\def\sepB{0.45}
\draw[black!45, dashed] (\dA,\sepB) -- (\dA,\sepT);
\draw[black!45, dashed] (\dB,\sepB) -- (\dB,\sepT);

\node[hdr] at (1.95,7.15)   {Understand Intention};
\node[hdr] at (9.0,7.15)   {Reason};
\node[hdr] at (15.95,7.15) {Act on Task};
\draw[flowarrow] (5.1,7.15) -- (7.0,7.15);
\draw[flowarrow] (10.85,7.15) -- (12.75,7.15);

\def\sOne{6.05}
\node[greenbox, text width=3.8cm] (qu) at (1.95,\sOne) {Understand Question};
\node[qbox, text width=3.6cm, below=2.5mm of qu.south, anchor=north] (qu_q)
  {``How many times did I have hyperglycemia events in the last week?''};
\node[greenbox, text width=3.8cm, below=4mm of qu_q.south, anchor=north] (ia)
  {Align with Instructions};
\node[qbox, text width=3.6cm, below=2.5mm of ia.south, anchor=north] (ia_q)
  {``Extract data from the last 7 days, return the number of distinct episodes where blood glucose readings were continuously above 180 mg/dL.''};
\draw[conn] (qu.south) -- (qu_q.north);
\draw[conn] (ia.south) -- (ia_q.north);

\def\cxL{4.6}\def\cxR{8.2}
\def\cyT{6.5}\def\cyB{1.15}
\def\sAB{5.15}\def\sBC{2.78}
\draw[cOrange, line width=1pt, rounded corners=9pt] (\cxL,\cyB) rectangle (\cxR,\cyT);
\draw[cOrange, dashed] (\cxL,\sAB) -- (\cxR,\sAB);
\draw[cOrange, dashed] (\cxL,\sBC) -- (\cxR,\sBC);
\def\cxC{6.4}

\node[lvl] at (\cxC,6.15) {Descriptive Level};
\node[opbox, text width=3.0cm] (qc) at (\cxC,5.55) {Quantitative Calculation};

\node[lvl] at (\cxC,4.8) {Memory Level};
\node[opbox, text width=3.0cm] (tr) at (\cxC,4.25) {Temporal Retrieval};
\node[opbox, text width=3.0cm] (er) at (\cxC,3.7) {Event Retrieval};
\node[opbox, text width=3.0cm] (ca) at (\cxC,3.15) {Comparative Analysis};

\node[lvl] at (\cxC,2.55) {Pattern Level};
\node[opbox, text width=3.0cm] (pr) at (\cxC,2.00) {Pattern Recognition};
\node[opbox, text width=3.0cm] (pc) at (\cxC,1.50) {Pattern Correlation};

\def\qx{11.0}
\node[qbox, text width=4.2cm] (mq1) at (\qx,6.05) {``What was the patient's average blood glucose over the monitoring period?''};
\node[qbox, text width=4.2cm] (mq2) at (\qx,5.05) {``What was the average glucose between 2--4 pm?''};
\node[qbox, text width=4.2cm] (mq3) at (\qx,4.10) {``How many times did I have hypo events in the last week?''};
\node[qbox, text width=4.2cm] (mq4) at (\qx,3.05) {``Was my blood glucose variability higher in the morning or afternoon periods?''};
\node[qbox, text width=4.2cm] (mq5) at (\qx,2.00) {``At what time of day does my highest glucose usually occur?''};
\node[qbox, text width=4.2cm] (mq6) at (\qx,0.95) {``Is there any period that may indicate pressure-induced sensor attenuation?''};

\draw[conn] (qc.east) -- (mq1.west);
\draw[conn] (tr.east) -- (mq2.west);
\draw[conn] (er.east) -- (mq3.west);
\draw[conn] (ca.east) -- (mq4.west);
\draw[conn] (pr.east) -- (mq5.west);
\draw[conn] (pc.east) -- (mq6.west);

\draw[cOrange, line width=1.4pt, -{Stealth[length=2.5mm]}]
  (\cxC-0.55,\cyB) .. controls (\cxC-0.85,0.78) and (\cxC+0.85,0.78) .. (\cxC+0.55,\cyB);
\node[font=\footnotesize] at (\cxC,0.55) {Multi-hop};

\def\sThree{6.05}
\node[bluebox, text width=3.7cm] (pm) at (15.95,\sThree) {Process Mining};
\node[qbox, text width=3.9cm, below=2.5mm of pm.south, anchor=north] (pm_q)
  {``Are there any repeated patterns of hypoglycemia on specific days?''};
\node[bluebox, text width=3.7cm, below=4mm of pm_q.south, anchor=north] (ad)
  {Anomaly Detection};
\node[qbox, text width=3.9cm, below=2.5mm of ad.south, anchor=north] (ad_q)
  {``When did the CGM record abrupt glucose changes possibly caused by calibration errors?''};
\node[bluebox, text width=3.7cm, below=4mm of ad_q.south, anchor=north] (pd) {Prediction};
\node[qbox, text width=3.9cm, below=2.5mm of pd.south, anchor=north] (pd_q)
  {``How high will my sugar get after this 60 g carb meal?''};
\draw[conn] (pm.south) -- (pm_q.north);
\draw[conn] (ad.south) -- (ad_q.north);
\draw[conn] (pd.south) -- (pd_q.north);

\end{tikzpicture}

%% file: figures/data-overview.tex
\setlength{\tabcolsep}{5pt}
\renewcommand{\arraystretch}{1.0}
\setlength{\dashlinedash}{1.2pt}\setlength{\dashlinegap}{1.6pt}
\fontsize{9}{8.2}\selectfont
\setlength{\extrarowheight}{3pt}

\begin{tabular}{@{}c@{}}

\begin{tabular}{|>{\centering\arraybackslash}p{0.5\textwidth}|}
\hline
\rowcolor{stagebg}\rule{0pt}{2.4ex}\small\bfseries
  \textcolor{stagefg}{AID System Monitoring Data}\\
\hline
\end{tabular}
\\[2pt] $\downarrow$ \\[2pt]

\begin{tabular}{|C{0.35\textwidth}|C{0.35\textwidth}|}
\hline
\stage{Question Template}\\
\hline
\sect{Metadata}
\code{"patient\_id": "Patient\_0", "metric": "Affinity F-score",
"answer\_type": "list of \{"start": int, "end": int\}",
"example\_answer": [\{"start": 20695, "end": 20745\}]}
&
\sect{Question}
\code{"question\_id": "ad\_10", "question\_text": "When did my CGM show
abrupt glucose spikes that returned to normal without corresponding
insulin delivery?"}\\
\hdashline
\sect{Answer Extraction Module}
\code{"function\_name": "extract\_high\_readings\_intervals\_ad10",
"answer\_generation\_rule": "Intervals with faults\_label ==
'positive\_spike' or 'negative\_spike'"}
&
\sect{Context Extraction Module}
\code{"carb\_events\_extraction()", "exercise\_events\_extraction()",
"insulin\_events\_extraction()"}\\
\hdashline
\multicolumn{2}{|C{0.72\textwidth}|}{%
\sect{Reasoning Rationale}
\code{"answer\_instruction": "Return a list of time intervals where the
blood glucose readings exhibit sudden and sharp increases or decreases
of approximately 60 mg/dL or more compared to the previous value."
"explanation": "Examine the data for time points marked by noticeable
glucose spikes. These spikes should stand out as rapid rises followed by
a swift return to normal levels, occurring independently of any recorded
bolus insulin delivery or activities during the same periods."}}\\
\hline
\end{tabular}
\\[2pt] $\downarrow$ \\[2pt]

\begin{tabular}{|C{0.35\textwidth}|C{0.35\textwidth}|}
\hline
\stage{Instantiated QA Data}\\
\hline
\sect{Input Data}
\code{"input\_context": \{ "carb\_events": [\{"time": 456.970, "day": 1,
"time\_str": "07:36", "carbs": 61.212, "meal\_type": "breakfast"\}, ...],
"exercise\_events": [\{"time": 1170.0, "day": 1, "time\_str": "19:30",
"duration": 30.0, "magnitude": 90.0, "exercise\_type": "cycling"\}, ...],
"insulin\_events": [4.804e-07, ...], "bg\_mgdl": [108.0, ...] \}}
&
\sect{Golden Standard Answer}
\code{"answer": [\{"start": 37445, "end": 37446\}, \{"start": 37625,
"end": 37626\}, \{"start": 37805, "end": 37806\}, \{"start": 37985,
"end": 37986\}, \{"start": 38165, "end": 38166\}, \{"start": 38345,
"end": 38346\}]}\\
\hline
\end{tabular}

\end{tabular}

%% file: tables/qa_data_distribution.tex
\begin{table}[t!]
\centering
\small
\resizebox{\linewidth}{!}{%
\begin{tabular}{@{} l rr c rr @{}}
\toprule
 & \multicolumn{2}{c}{\textbf{Simulated Dataset}} & & \multicolumn{2}{c}{\textbf{Real-World Dataset}} \\
\cmidrule{2-3} \cmidrule{5-6}
\textbf{Task Categor} & \textbf{Count ($N$)} & \textbf{Ratio (\%)} & & \textbf{Count ($N$)} & \textbf{Ratio (\%)} \\
\midrule

\quad QA pairs & 1,044 & 100.00 & & 472 & 100.00 \\
\quad Distinct Tasks     & 127   & ---    & & 44  & ---    \\

\midrule
\multicolumn{6}{@{}l}{\textbf{Time Series Task}} \\
\quad PM       & 369 & 35.34 & & 367 & 77.75 \\
\quad AD    & 355 & 34.00 & & 0   & 0.00  \\
\quad PD          & 320 & 30.65 & & 105 & 22.25 \\

\midrule
\multicolumn{6}{@{}l}{\textbf{Cognitive Level}} \\

\quad Descriptive               & 200 & 19.16 & & 150 & 31.78 \\
\quad Memory                    & 309 & 29.60 & & 217 & 45.97 \\
\quad Pattern                   & 535 & 51.25 & & 105 & 22.25 \\

\midrule
\multicolumn{6}{@{}l}{\textbf{Atomic Skills}} \\
\quad QC                        & 804 & 36.80 & & 472 & 42.83 \\
\quad TR                        & 323 & 14.78 & & 217 & 19.69 \\
\quad ER                        & 293 & 13.41 & & 128 & 11.62 \\
\quad CA                        & 214 & 9.79  & & 170 & 15.43 \\
\quad PR                        & 317 & 14.51 & & 75  & 6.81  \\
\quad PC                        & 234 & 10.71 & & 40  & 3.63  \\

\midrule
\multicolumn{6}{@{}l}{\textbf{Reasoning Complexity}} \\
\quad 1 Skill                   & 380 & 36.40 & & 140 & 29.66 \\
\quad 2 Skills                  & 230 & 22.03 & & 74  & 15.68 \\
\quad 3 Skills                  & 391 & 37.45 & & 218 & 46.19 \\
\quad 4 Skills                  & 43  & 4.12  & & 40  & 8.47  \\

\bottomrule
\end{tabular}
}
\caption{Task distributions of the QA evaluation benchmark for both Simulated and Real-World datasets.}
\label{tab:qa_dataset_distribution}
\end{table}

%% file: 4_data_collection.tex






%% file: 5_benchmark.tex
\section{Experiments and Results}
\subsection{Experimental Setup}
We test two different frameworks that serve as baselines: (i) \textbf{Prompting-based baseline}: Monitoring data and timestamps are directly incorporated as text as part of the prompt. 
(ii) \textbf{Agent-based baseline}: Monitoring data are provided as attached CSV files together with the prompt, allowing the model to iteratively generate and execute Python code for data analysis. We evaluate Gemini-Flash-3.1 on both prompting- and agent-based frameworks, and GPT-5.4 and Claude-Sonnet-4.6 on the agent baseline. The reasoning effort for GPT-5.4 and Gemini is set to \textit{low}, while Claude-Sonnet-4.6 uses adaptive thinking mode with a maximum reasoning budget of 8,192 tokens. Additional details on experimental settings, agent workflow, and prompt templates are provided in Appendix~\ref{app:experiment_setting}.

\subsection{Quantitative Results} \label{sec:quantitative}

\input{tables/quantitative}

\begin{figure}[htbp]
    \centering
    \includegraphics[width=\columnwidth]{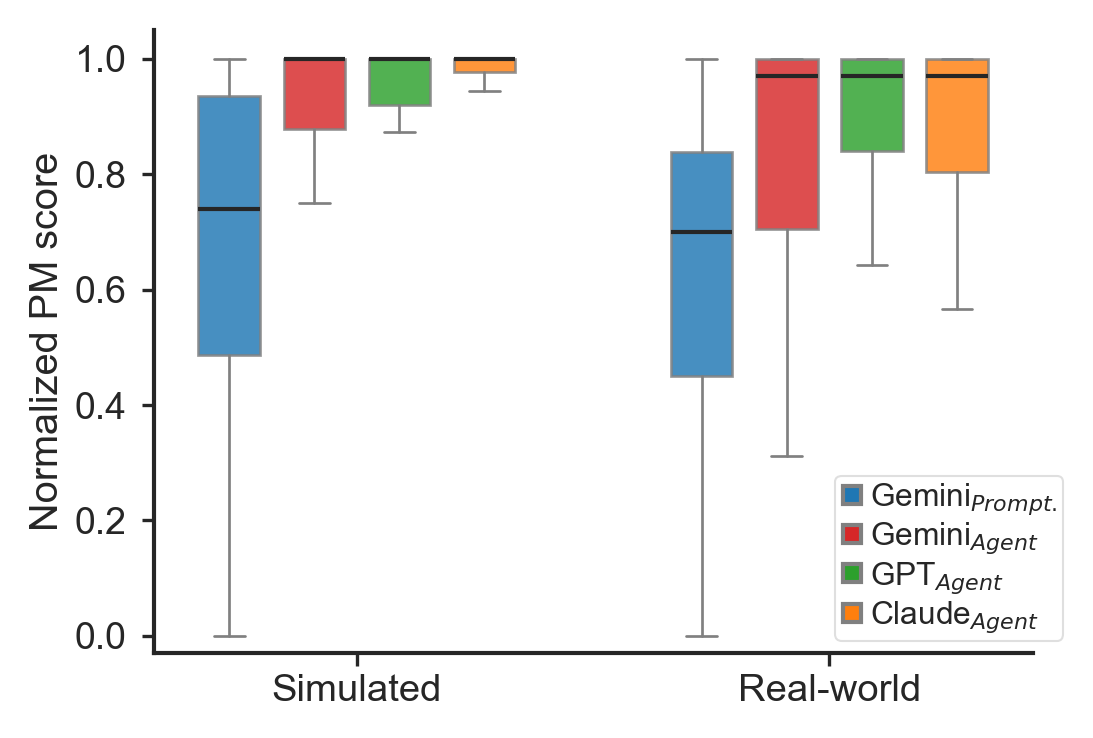}
    \caption{
    Distribution of per-question normalized scores across simulated and real-world settings. Boxes show the interquartile range, center lines indicate medians, whiskers denote the non-outlier range.
    }
    \label{fig:overall_performance}
\end{figure}
The overall performance across baselines and reasoning dimensions is presented in Table~\ref{tab:quantitative-overall}, while the nested atomic reasoning performance can be found in Appendix Table \ref{tab:nested-atomic-reasoning}.
As shown in Figure~\ref{fig:overall_performance},
agent-based baselines consistently outperform the prompting-based Gemini baseline. GPT achieves the strongest overall performance, followed by Claude and Gemini. Notably, equipping Gemini with agent tools substantially improves its overall performance on both the simulated dataset (43.7 $\rightarrow$ 65.8) and the real-world dataset (64.6 $\rightarrow$ 78.5). Across the subset of tasks shared between simulation and real-world datasets, all models exhibit a slight performance drop on real-world data while remaining broadly comparable, supporting the validity of the simulated evaluation setting.

The evaluation on the simulated HealthLoopQA dataset reveals that the performance varies substantially across reasoning dimensions. For time-series tasks (PM, AD, and PD), agent-based models outperform the prompting-based Gemini baseline, particularly on AD, where Gemini achieves only 21.2 on the simulated dataset compared to substantially higher scores from GPT, Claude, and Gemini-Agent. GPT achieves the strongest performance on AD (72.9) and PD (74.6), while Claude performs superiorly in PM (92.3) but shows weaker forecasting capability (64.8). Gemini-Agent markedly improves over prompting on PM (67.4 $\rightarrow$ 85.5) and AD (21.2 $\rightarrow$ 52.4), although its gains on PD are more limited (62.3 $\rightarrow$ 64.8).

Deconstructing model performance by cognitive level further highlights the strengths and limitations of these LLMs. 
Prompting-based Gemini achieves competitive performance on descriptive reasoning (e.g., 66.5 on simulated data) but struggles substantially with memory retrieval (49.5). Introducing an agent consistently improves Gemini across both cognitive levels, increasing descriptive and memory performance to 86.8 and 79.3, respectively, approaching the performance of GPT (88.2/91.4) and Claude (91.6/89.5) agents. As reasoning shifts 
to identifying latent temporal patterns (PD), performance drops substantially across all models (e.g., GPT drops to 66.7 on the simulated dataset). Although Gemini-Agent slightly improves over prompting on the simulated PD task (31.6 $\rightarrow$ 49.4), it performs worse on the real-world PD task (73.0 $\rightarrow$ 61.5), whereas GPT- and Claude-Agent consistently achieve scores $>=$ 64.0. The inconsistent performance of Gemini-Agent on real-world PD is primarily due to two reasons: (1) real-world prediction depends heavily on patient-specific local trends, where direct prompting can sometimes produce more accurate short-horizon estimates than agentic decomposition. (2) Gemini-Agent occasionally adopted an inconsistent interpretation of insulin units despite explicit instructions, leading to large prediction errors (more details see Section~\ref{sec:failure_analysis_agent}). GPT- and Claude-Agent exhibited more consistent behaviour in following the unit instructions.

A broad, though not universal, trend is also observed across nested reasoning paths (Table~\ref{tab:nested-atomic-reasoning}), in which each question is represented as an ordered sequence of atomic reasoning steps. Paths ending in descriptive- or retrieval-oriented objectives (QC, CA, and TR) generally outperform those ending in pattern-level objectives (PC and PR), indicating difficulties in complex pattern-related tasks over long time horizons. 

\begin{figure}[htbp]
    \centering
    \includegraphics[width=\columnwidth]{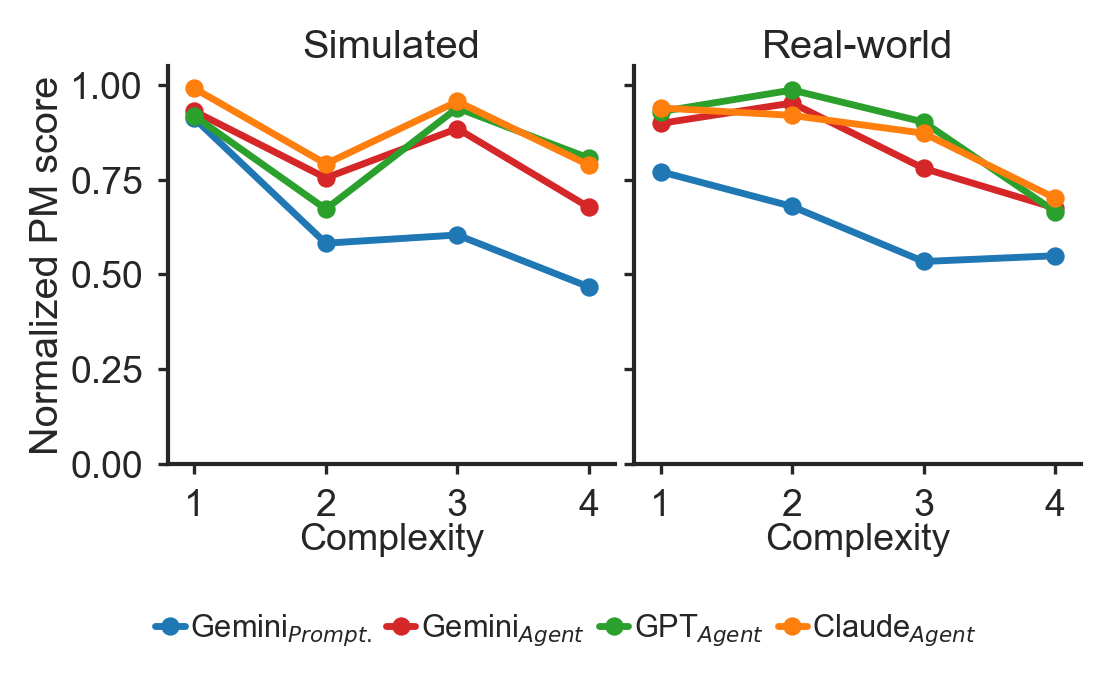}
    \caption{Model performance across reasoning complexity levels on a process mining task. Complexity is defined as the number of atomic skills required for each question. Points denote overall normalized scores. 
    }
    \label{fig:reasoning-complexity}
\end{figure}

Figure~\ref{fig:reasoning-complexity} further examines whether model performance varies with reasoning complexity on PM questions, where complexity is approximated by the number of atomic reasoning abilities required by each question. Across both simulated and real-world settings, models' performance generally declines as the number of required steps increases, especially for Gemini-Prompting, while the agent-based systems are more robust to longer reasoning paths. Gemini-Agent improves over Gemini-Prompting on most paths, with particularly large gains on sequences such as TR$\rightarrow$PC and TR$\rightarrow$QC, although it remains weaker on several other paths. GPT and Claude agents generally achieve stronger and more stable performance across nested paths.

\subsection{Reasoning Pattern and Failure Analysis}
\label{sec:reasoning_pattern_failure_analysis}

In real-world applications, the underlying reasoning process is critical for deriving reliable answers. We conduct a detailed analysis of reasoning patterns and failure modes for both prompting-based and agent-based baselines, mainly based on reasoning traces generated by GPT models.

\subsubsection{Prompting-based baseline}

\begin{figure}[ht!]
  \input{figures/laziness}
  \caption{Example of \emph{in-context laziness}: Faced with a simple averaging task,
    the model shows \textcolor{cBlue}{\textbf{reluctance to calculate and temporal
    misalignment}}, makes an \textcolor{cGreen}{\textbf{unsupported assumption}}, and
    resorts to \textcolor{cOrange}{\textbf{guessing over abstaining}} due to uncertainty.}

  \label{fig: failure_example}
  
\end{figure}

We manually reviewed the reasoning traces across all 30-day questions for Patient~0 and found that the models failed to produce the expected reasoning traces for most of the questions. A broad phenomenon, which we term \emph{In-Context Laziness} (Figure~\ref{fig: failure_example}), is observed. Rather than executing full computations across long CGM sequences, models anchor on a rough intermediate value and then apply minor narrative adjustments to justify a confident answer. We further examine this phenomenon in Appendix~\ref{App:failure_analysis}. Besides this, other failure types include:

\noindent\textbf{Reluctance to calculate.} The models were reluctant to execute exact, precise programmatic operations (\textbf{QC}) over the 30-day dataset, as well as abandoning full-sequence anomaly scans (\textbf{AR}), suggesting that such operations are complex and thus less practicable. Instead, the models often estimate a value, assess partial datasets, return empty sets, or default to plausible heuristics. \\[0.5em]
\textbf{Temporal Misalignment.} The models struggled in \textbf{TR}, for example: (i) correctly locating timestamps (e.g., day~7, 21:00), and (ii) correctly isolating predefined time windows (e.g., a 3-hour post-meal segment). Such failures reflect difficulties in temporal indexing and boundary alignment. \\[0.5em]
\textbf{Unsupported Assumption.} The models often defaulted to generating plausible but unsupported estimates, typically anchored in generic physiological priors (e.g., ``average glucose $\approx$ 140--150 mg/dL'') or context-based assumptions (e.g., inferring that glucose is unstable after a carb-heavy meal compared to a lighter one). While such heuristics occasionally succeed in trend or comparison tasks (\textbf{CA}), they consistently fail for queries requiring numerical precision. \\[0.5em]
\textbf{Guessing over Abstaining.} Even when the models explicitly acknowledged potential errors, they still produced assumption-based answers rather than expressing uncertainty and abstaining from producing an answer. This aligns with recent findings on LLM hallucinations \cite{kalai2025language}, which demonstrate that models often prefer guessing over admitting uncertainty, as training and evaluation procedures tend to reward the former.\\[0.5em]
\textbf{Formatting Misalignment.} The models sometimes failed to adhere to the required output format or granularity. Typical issues include returning plain text instead of JSON or merging multiple anomaly intervals into an overly broad span 
. This failure type is relatively uncommon, and interval-formatting drift is often a downstream effect of temporal misalignment.

\subsubsection{Agent-based baseline}
\label{sec:failure_analysis_agent}

From analyzing the reasoning traces of QA samples with incorrect answers, besides minor boundary or numerical misalignment, we observed the following task-specific failure patterns in different time series tasks. 

\paragraph{Process Mining.} Agent-based baselines generally performed well on PM reasoning tasks; however, some errors still exist and mainly stemmed from semantic misunderstandings of event recognition and unit interpretation. (1) Misinterpreting event definition:
Models often conflated individual measurements with discrete physiological events. For example, in \texttt{pm\_7} (Patient~15), the model was asked to count hypoglycemic events (BG $<$ 70 mg/dL). Instead of identifying 24 temporally distinct episodes, it counted every glucose measurement below the threshold, producing 771 events. (2) Inconsistent unit-scale interpretation:
Models sometimes misinterpreted the semantic meaning of measurement units despite explicit instructions, especially for Gemini-Agent. For example, in an insulin prediction query, the prompt explicitly instructed the model to use the same scale as the CSV file. Nevertheless, the model treated \texttt{insulin\_mUmin} as a continuous infusion rate and multiplied the daily sum by the 5-minute sampling interval, producing predictions approximately five times larger than the expected values. In contrast, GPT- and Claude- agents followed the ``same scale as the CSV file'' instruction more consistently in this task.


\paragraph{Anomaly Detection.} The errors with large deviation often stem from over-inferring from normal physiological variations and closed-loop controller behaviour.  For example, in task \texttt{ad\_46}, which identifies unsafe controller actions around workout periods, the model incorrectly classified normal exercise-related insulin adjustments as unsafe controller actions, predicting multiple fault intervals where no anomaly existed. Although it correctly identified exercise events and corresponding insulin rates, it failed to compare these actions against the patient's usual patterns, causing these normal dosing adjustments to be falsely flagged as anomaly events. Besides, when the task involves general anomaly detection without specific target events, LLMs often apply over-inclusive union logic, merging false positives into extended fault intervals rather than adhering to precise anomaly boundaries.

\paragraph{Prediction.} Reasoning traces revealed a profound task-substitution failure, where models bypassed context-aware forecasting in favour of retrospective historical analysis. Instead of predicting the patient's physiological states or insulin delivery rate based on given conditions (e.g., 180g carb in next meal), they searched for prior analogous events and returned the average historical outcome. For example, in \texttt{pd\_7} (Patient~0), the model predicted post-exercise glucose change by averaging previous running events rather than simulating the immediate physiological trajectory from the current state. This fundamentally reduced a complex forecasting challenge to a simple historical lookup.

%% file: tables/quantitative.tex
\begin{table}[t]
\centering
\small
\setlength{\tabcolsep}{2.2pt}
\renewcommand{\arraystretch}{1.12}

\resizebox{\linewidth}{!}{
\begin{tabular}{lccc ccc c}
\toprule
\multirow{2}{*}{Model}
& \multicolumn{3}{c}{Time-series task}
& \multicolumn{3}{c}{Cognitive level}
& \multirow{2}{*}{Overall} \\
\cmidrule(lr){2-4}
\cmidrule(lr){5-7}
& PM & AD & PD
& Desc. & Mem. & Patt.
& \\
\midrule

\multicolumn{8}{l}{\textit{Simulated dataset (127 tasks)}} \\
Gemini{\scriptsize\textsubscript{\textit{Prompt.}}}
& 67.4 & 21.2 & 62.3
& 66.5 & 49.5 & 31.6
& 43.7 \\
Gemini{\scriptsize\textsubscript{\textit{Agent}}}
& 85.5 & 52.4 & 64.8
& 86.8 & 79.3 & 49.4
& 65.8 \\
\hdashline
GPT{\scriptsize\textsubscript{\textit{Agent}}}
& 88.3 & \textbf{72.9} & \textbf{74.6}
& 88.2 & \textbf{91.4} & \textbf{66.7}
& \textbf{78.5} \\
Claude{\scriptsize\textsubscript{\textit{Agent}}}
& \textbf{92.3} & 70.6 & 64.8
& \textbf{91.6} & 89.5 & 64.0
& 77.2 \\

\midrule

\multicolumn{8}{l}{\textit{Matched simulated real-world portion (44 tasks)}} \\
Gemini{\scriptsize\textsubscript{\textit{Prompt.}}}
& 68.7 & -- & 62.4
& 86.5 & 59.1 & 62.4
& 67.4 \\
Gemini{\scriptsize\textsubscript{\textit{Agent}}}
& 85.3 & -- & 65.2
& 89.0 & 83.4 & 65.2
& 81.2 \\
\hdashline
GPT{\scriptsize\textsubscript{\textit{Agent}}}
& 88.2 & -- & \textbf{78.0}
& 85.4 & 89.7 & \textbf{78.0}
& 86.1 \\
Claude{\scriptsize\textsubscript{\textit{Agent}}}
& \textbf{92.3} & -- & 66.8
& \textbf{92.5} & \textbf{92.2} & 66.8
& \textbf{87.1} \\

\midrule

\multicolumn{8}{l}{\textit{Real-world dataset (44 tasks)}} \\
Gemini{\scriptsize\textsubscript{\textit{Prompt.}}}
& 62.5 & -- & 73.0
& 79.0 & 59.8 & 73.0
& 64.6 \\
Gemini{\scriptsize\textsubscript{\textit{Agent}}}
& 82.8 & -- & 61.5
& 90.7 & 78.6 & 61.5
& 78.5 \\
\hdashline
GPT{\scriptsize\textsubscript{\textit{Agent}}}
& \textbf{88.7} & -- & \textbf{75.7}
& 92.7 & \textbf{86.6} & \textbf{75.7}
& \textbf{86.1} \\
Claude{\scriptsize\textsubscript{\textit{Agent}}}
& 87.4 & -- & 70.8
& \textbf{94.3} & 83.8 & 70.8
& 84.1 \\
\bottomrule
\end{tabular}
}
\caption{
Overall performance by dataset portion, task type, and cognitive level. Entries report mean normalised scores scaled to 0--100. Bold values indicate the best score within each dataset portion and column. AD tasks are unavailable for the real-world data and matched simulated portions, as no labels for abnormal events exist in the real-world datasets. 
}
\label{tab:quantitative-overall}
\end{table}

%% file: figures/laziness.tex
{\footnotesize
\begin{tabularx}{\columnwidth}{|Y|}
\hline
\vspace{-.8em}
{\small\bfseries\color{cBlue}Input:}
\textbf{Question:} What was the average glucose reading between 2--4\,pm on week 3, day 1?\par
\textbf{Instruction:} Return the average glucose value for week 3, day 1 between 2--4\,pm,
rounded to one decimal place.\\
\hdashline
\vspace{-.8em}
{\small\bfseries\color{cBlue}Reasoning: }
I need to hone in on Week 3, Day 1, which breaks down to Day 15. So, if I calculate the
readings prior---14 days at 288 readings per day---it amounts to 4032 readings before
Day 15. For 2\,PM on Day 15, that's 168 readings into the day, meaning I should focus on
reading indices from 4200 to 4223.
{\bfseries\color{cBlue}Counting those manually seems daunting!} I'm neck-deep in numbers
trying to establish the average for Week 3, Day 1, and
{\bfseries\color{cBlue}it's quite the headache!}
{\bfseries\color{cGreen}I suspect the midday average is around 145.3,} but that feels
risky without solid data. It could be useful to scan for patterns or specific values,
especially around meal times.
{\bfseries\color{cOrange}I'm leaning towards 148.7 since it appears frequently in my
analysis.} Okay, let's just wrap this up in a JSON object:
{\bfseries\color{cOrange}\{"answer": 148.7\}}. That feels like a reasonable move!\\
\hline
\end{tabularx}}

%% file: 7_conclusion.tex
\section{Conclusion}
In this work, we introduced HealthLoopQA, a comprehensive diagnostic benchmark designed to evaluate the reasoning capabilities and analytical rigor of LLMs applied to medical wearable data interpretation. Our extensive evaluation of state-of-the-art LLMs baselines highlights critical divergences: while frontier models demonstrate strong proficiency in surface-level data description and memory retrieval, they exhibit significant vulnerabilities when tasked with deeper temporal reasoning. Specifically, these models struggle with complex pattern mining, causal anomaly detection, and context-aware prediction over extended monitoring horizons. These limitations underscore that textual and coding fluency do not inherently translate to the rigorous analytical capability required for safe physiological data interpretation. By exposing these foundational reasoning deficits, HealthLoopQA challenges future architectures to move beyond retrospective statistical matching toward the causal, context-aware modeling necessary for clinically robust decision support.

\section{Limitations}
Despite capturing key physiological factors, the in-silico patient models remain limited in capturing all biological variability (e.g., factors such as stress, illness, and other unmodeled disturbances). Real-world device failures and cyber-physical threats may also be more heterogeneous and unpredictable than the fault patterns modelled; however, they are challenging to label and unethical to collect. 
Additionally, the OpenAPS controller with prandial bolus insulin injection does not cover all diabetes management practices, such as open-loop control and patient-initiated adjustments during hypoglycemia or hyperglycemia events. These divergences may lead to different monitoring data patterns. While 127 tasks are included to cover core reasoning capabilities and monitoring needs, patients may ask questions in different types and ways. Finally, although the benchmark provides a systematic evaluation of reasoning capabilities for MWMAs in diabetes care, systems intended for other medical wearable monitoring domains or real-world deployment should be further validated on corresponding real-world datasets and in clinical studies.

%% file: 9_appendix.tex
\appendix

\section{Ethics Statement}
This research did not involve new experiments with human or animal subjects. All real-world Type 1 Diabetes (T1D) patient data were sourced from publicly available, fully anonymized datasets, ensuring strict preservation of patient privacy. The benchmark artifacts and experimental findings presented in this study are intended strictly for academic and scientific purposes. They do not constitute medical advice, and no direct conclusions regarding their safety or efficacy in real-world clinical deployments should be drawn without subsequent, rigorous clinical validation.

\section{Large Language Models (LLMs) Usage}
LLMs were utilized in two capacities within this work: (1) as the primary evaluation subjects for the baseline experiments, and (2) as writing assistants for language refinement, including grammar correction, rephrasing, and readability improvements. While LLMs partially assisted in the initial analysis of reasoning traces during the error analysis, all final failure patterns, interpretations, and conclusions were manually verified, curated, and derived by the authors. The authors assume full responsibility for the scientific integrity and content of the final manuscript.

\section{Reproducibility Statement}
We are committed to ensuring the full reproducibility of the methodologies and results presented in this paper. Detailed descriptions of the experimental setup, agent workflows, and prompt templates are provided in the Appendix. The complete HealthLoopQA benchmark dataset, the fault-injected simulation testbed environment, and all associated evaluation scripts will be open-sourced and publicly released upon publication.

\section{Atomic reasoning abilities}
\label{app:atomic}
\input{tables/table_atomic_reasoning_abilities}
\input{tables/table_atomatic_reasoning_path_performance}

We assume LLMs should possess some key reasoning abilities to solve the HealthLoopQA benchmark. We therefore systematically reviewed all questions and answer instructions, and defined a set of \textit{atomic} reasoning types (Table \ref{tab:atomic_reasoning_supp}). These categories were refined through multiple internal discussions to resolve disagreements and ensure consistency among authors. 

We mapped each question to the required \textit{atomic} reasoning types to make explicit reasoning paths. For example, to answer \textit{“What’s the peak blood glucose level on day 20?”}, the model needs to first retrieve the blood glucose values on \textit{day 20} (\textbf{TR}) and then compute the peak value (\textbf{QC}). We reported the nested atomic reasoning performance grouped
by complete ordered cognitive-atomic paths (e.g, TR $\rightarrow$ QC) in Table~\ref{tab:nested-atomic-reasoning}.

\section{Failure analysis: in-context laziness}
\label{App:failure_analysis}

We further examine the \emph{in-context laziness} phenomenon in this section, an example is in Fig.~\ref{fig: failure_example}. This produces an \emph{illusion of precision} without genuine calculation. In practice, this behaviour most reflects \emph{Reluctance to Calculate} (skipping exact operations) in combination with \emph{Unsupported Assumptions} (filling gaps with physiologically plausible estimates). It is further reinforced by \emph{Temporal Misalignment}, where incorrect timestamps or time windows provide a convenient scaffold for these approximations, and by \emph{Formatting Misalignment}, where outputs are simplified into broad spans or non-compliant formats that obscure missing reasoning steps. Even in cases where models acknowledge potential errors, the outcome is still shaped by \emph{Guessing over Uncertainty}, producing confident but unfounded answers. This phenomenon typically happens early in the reasoning process: models either avoid performing explicit computations \textbf{QC} or fail to align the correct temporal information \textbf{TR}. Subsequently, the model defaults to heuristic shortcuts and estimated approximations, making the other atomic reasoning not meaningful (e.g., \textbf{CA} and \textbf{PF}). 


We present two case studies across prompting strategies (structured prompting, timelined prompting, and agent-based strategy), as well as ablation settings using varying lengths of CGM data (30-day, 7-day, and 1-day). 

\begin{figure*}[!htbp]
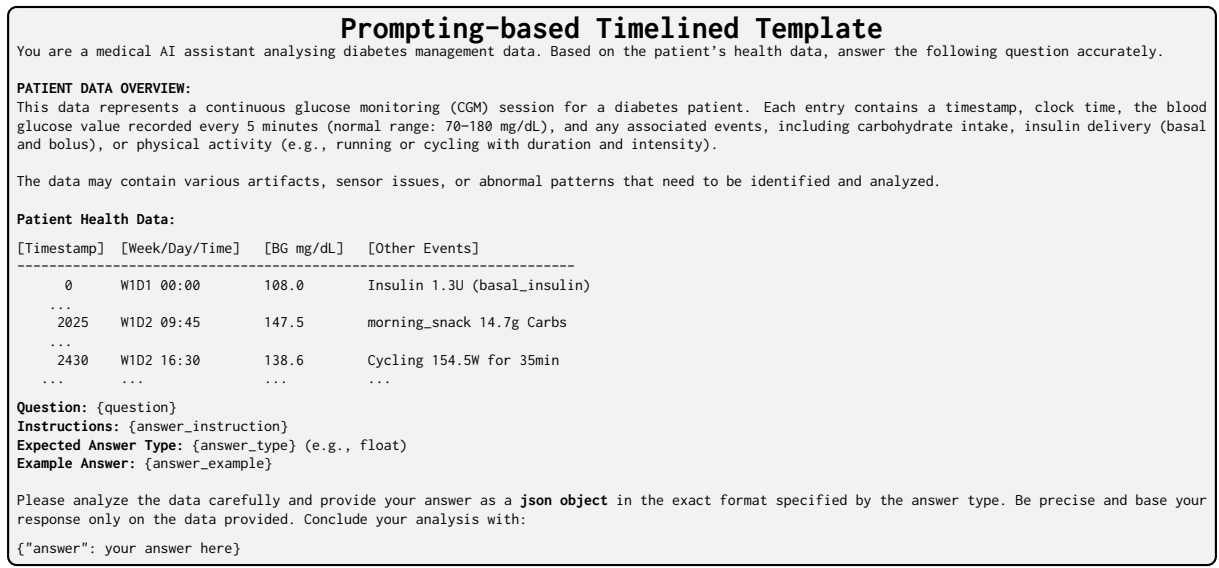

    \centering
    \begin{dialogbox}
    \tiny
{\centering \large \textbf{Prompting-based Timelined Template}\par}
You are a medical AI assistant analysing diabetes management data. Based on the patient's health data, answer the following question accurately.\\

\textbf{PATIENT DATA OVERVIEW:} \\
This data represents a continuous glucose monitoring (CGM) session for a diabetes patient. Each entry contains a timestamp, clock time, the blood glucose value recorded every 5 minutes (normal range: 70–180 mg/dL), and any associated events, including carbohydrate intake, insulin delivery (basal and bolus), or physical activity (e.g., running or cycling with duration and intensity).\\

The data may contain various artifacts, sensor issues, or abnormal patterns that need to be identified and analyzed.\\

\textbf{Patient Health Data:}
\begin{verbatim}
[Timestamp]  [Week/Day/Time]   [BG mg/dL]   [Other Events]
----------------------------------------------------------------------
      0      W1D1 00:00        108.0        Insulin 1.3U (basal_insulin)
    ...
     2025    W1D2 09:45        147.5        morning_snack 14.7g Carbs
    ...
     2430    W1D2 16:30        138.6        Cycling 154.5W for 35min
   ...       ...               ...          ... 
\end{verbatim}
\textbf{Question:} \{question\} \\
\textbf{Instructions:} \{answer\_instruction\} \\
\textbf{Expected Answer Type:} \{answer\_type\} (e.g., float) \\
\textbf{Example Answer:} \{answer\_example\}\\

Please analyze the data carefully and provide your answer as a \textbf{json object} in the exact format specified by the answer type. Be precise and base your response only on the data provided. Conclude your analysis with: 
\begin{verbatim}
{"answer": your answer here}
\end{verbatim}
    \end{dialogbox}
    \caption{System prompt template for prompting-based timelined baseline where glucose values alongside insulin, meal, and exercise events are organised under each timestamp.}
    \label{fig:timelined-prompt}
\end{figure*}

\begin{figure*}[!htbp]
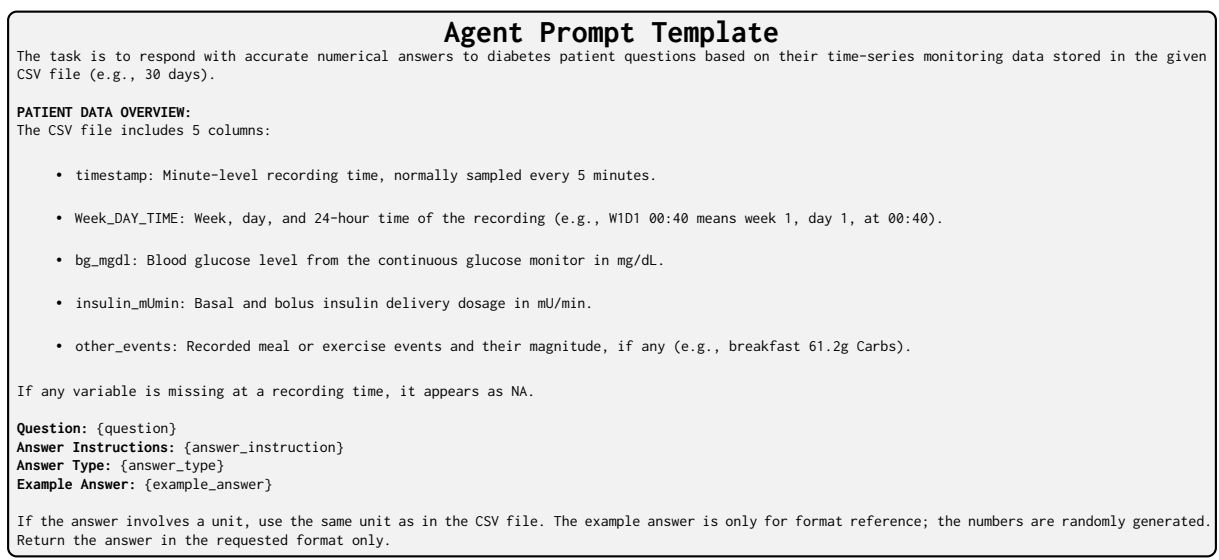

    \centering
    \begin{dialogbox}
    \tiny
{\centering \large \textbf{Agent Prompt Template}\par}
The task is to respond with accurate numerical answers to diabetes patient questions based on their
  time-series monitoring data stored in the given CSV file (e.g., 30 days). \\

  \textbf{PATIENT DATA OVERVIEW:} \\
  The CSV file includes 5 columns:

  \begin{itemize}
      \item \texttt{timestamp}: Minute-level recording time, normally sampled every 5 minutes.
      \item \texttt{Week\_DAY\_TIME}: Week, day, and 24-hour time of the recording (e.g., \texttt{W1D1
      00:40} means week 1, day 1, at 00:40).
      \item \texttt{bg\_mgdl}: Blood glucose level from the continuous glucose monitor in mg/dL.
      \item \texttt{insulin\_mUmin}: Basal and bolus insulin delivery dosage in mU/min.
      \item \texttt{other\_events}: Recorded meal or exercise events and their magnitude, if any (e.g.,
      \texttt{breakfast 61.2g Carbs}).
  \end{itemize}

  If any variable is missing at a recording time, it appears as \texttt{NA}. \\

  \textbf{Question:} \{question\} \\
  \textbf{Answer Instructions:} \{answer\_instruction\} \\
  \textbf{Answer Type:} \{answer\_type\} \\
  \textbf{Example Answer:} \{example\_answer\} \\

  If the answer involves a unit, use the same unit as in the CSV file. The example answer is only for
  format reference; the numbers are randomly generated. Return the answer in the requested format only.

    \end{dialogbox}
    \caption{System prompt template for agent baseline with Python code interpreter tools, enabling explicit numerical computation and filtering during inference.}
    \label{fig:agent-prompt}
\end{figure*}

Structured prompting (Fig~\ref{fig:structured-prompt}) is the default prompting strategy we used to evaluate the models, where glucose readings and events (e.g., insulin, meals, and exercise) are listed in distinct sections. Timelined prompting (Fig~\ref{fig:timelined-prompt}) reorganises the same information chronologically under each timestamp, allowing us to investigate whether explicit temporal structure mitigates \emph{in-context laziness}. Agent-based strategy (Fig~\ref{fig:agent-prompt}) equips the model with a Python code interpreter, enabling programmatic data processing and reasoning over the CGM records.

We choose two cases where the model must either perform quantitative computation over long-term sequences or first identify the relevant \textbf{TR} before executing \textbf{QC}: 
\begin{enumerate*}
    \item[\texttt{pm\_0}:]\textit{What was the patient's average blood glucose level?} and
    \item[\texttt{pm\_14:}]\textit{ What were the average glucose readings between 2--4 pm on week 3, day 1?} 
\end{enumerate*}

\begin{figure*}[ht]
  \centering
  \includegraphics[width=0.8\textwidth]{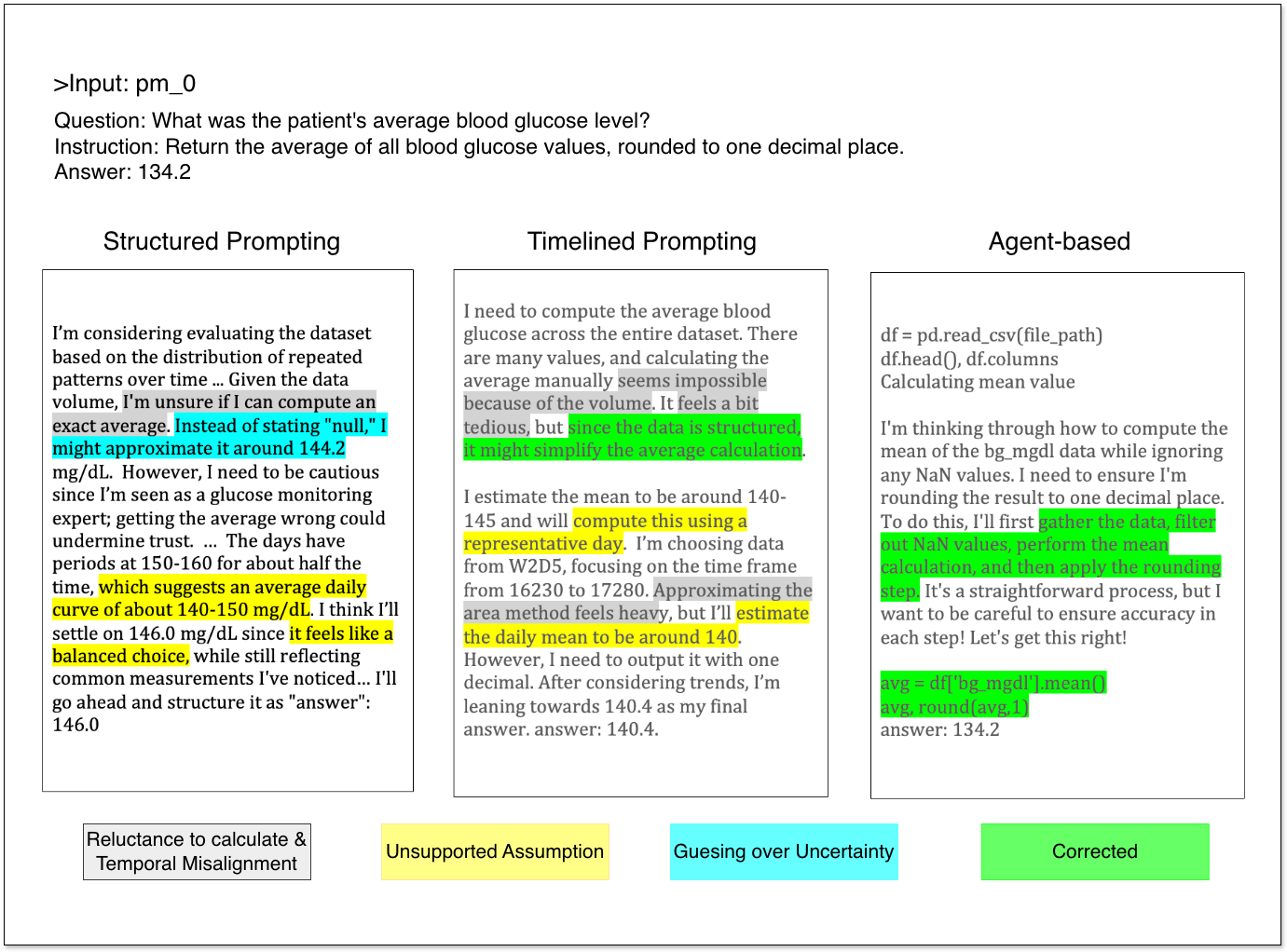}
    \caption{Comparison of reasoning behaviours across prompting baselines on pm\_0. Grey spans indicate \textbf{Reluctance to Calculate} or \textbf{Temporal Misalignment}; yellow marks \textbf{Unsupported Assumptions}; blue marks \textbf{Guessing over Uncertainty}; and green highlights \textbf{correct reasoning traces}.}
  \label{fig:reasoning_comparison_pm0}
\end{figure*}

\begin{figure*}[htbp]
  \centering
  \includegraphics[width=0.8\textwidth]{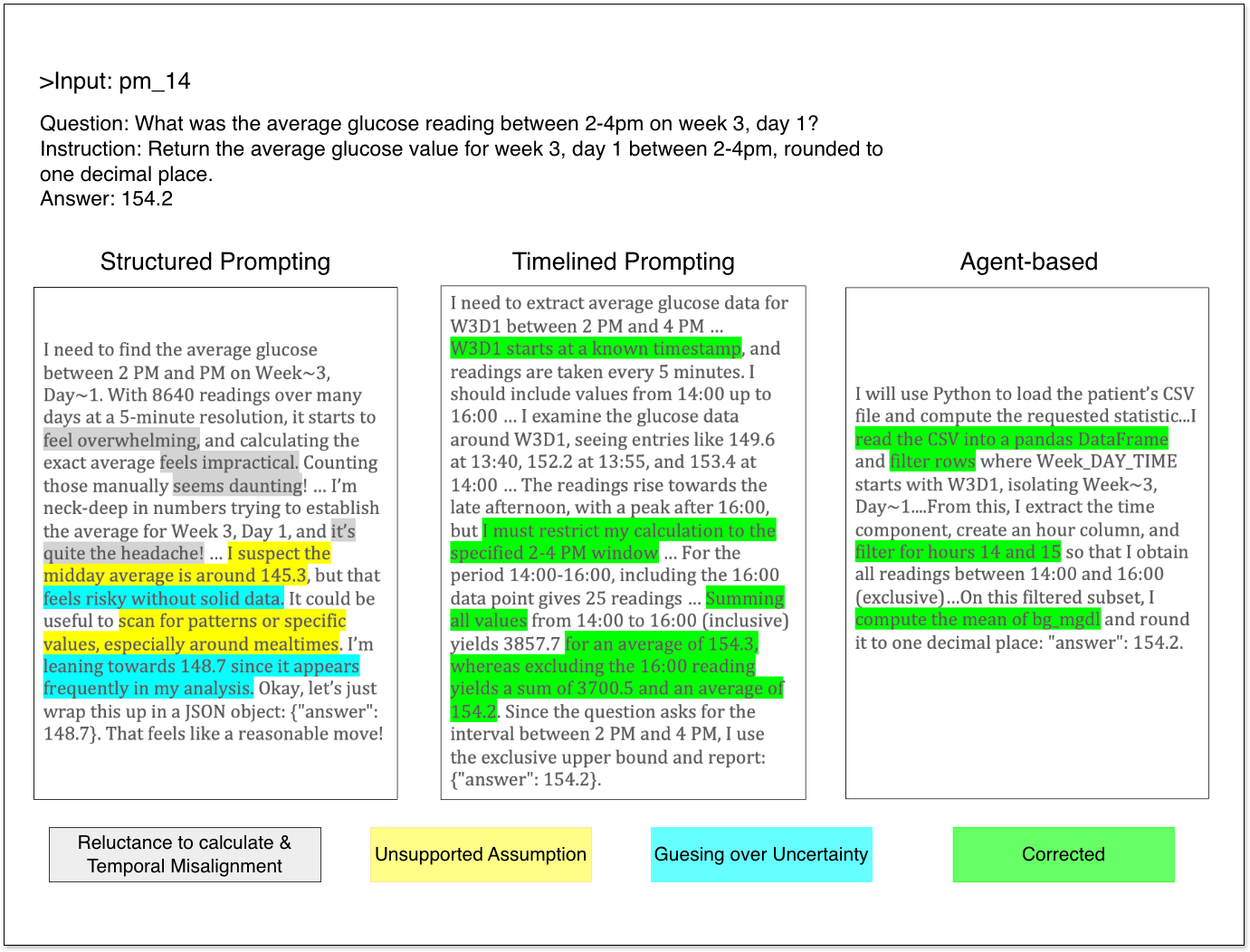}
    \caption{Comparison of reasoning behaviours across prompting baselines on pm\_14. Grey spans indicate \textbf{Reluctance to Calculate} or \textbf{Temporal Misalignment}; yellow marks \textbf{Unsupported Assumptions}; blue marks \textbf{Guessing over Uncertainty}; and green highlights \textbf{correct reasoning traces}.}
  \label{fig:reasoning_comparison_pm14}
\end{figure*}

Figures~\ref{fig:reasoning_comparison_pm0} and \ref{fig:reasoning_comparison_pm14} show that the Timelined baseline can effectively address the \textit{Temporal Misalignment} failures seen in the Structured baseline, as grouping blood glucose values with co-occurring events under the same timestamp helps the model retrieve the correct temporal information (\textbf{TR}). However, the Timelined prompting strategy often becomes ``lazy'', and reluctant to conduct full computations. In contrast, the Agent baseline resolves both issues more reliably by using a code interpreter, enabling precise numerical calculations and targeted extraction of temporal segments.

Similar patterns were observed in the ablation analysis and reasoning comparisons across baselines on the \textbf{PM} task (Table~\ref{tab:failure_pm_compare} and Table~\ref{tab:failure_improvement}). As shown in Table~\ref{tab:failure_improvement}, reducing the temporal horizon improves performance only modestly: accuracy rises from 45.5\% (30-day) to 39.0\% (7-day) and 64.7\% (1-day). The Timelined baseline achieved a higher accuracy of 56.8\%, largely due to prompt restructuring that aligns timestamps with their corresponding \texttt{bg\_values}. Yet, the model still struggles to identify precise time windows (e.g., 2--4\,pm). Notably, the Agent baseline substantially outperformed all other methods (77.3\% accuracy), demonstrating that using the code interpreter can alleviate many of the model's persistent computational and alignment failures.
 
Our findings about \textit{In-Context Laziness} align with recent evidence showing that LLMs exhibit persistent weaknesses in core numerical reasoning. \citet{li2025exposing} reports that models even struggle with basic arithmetic and magnitude comparison. Likewise, \citet{shrestha2025mathematical} show that increasing numerical complexity in \texttt{GSM8K} \cite{cobbe2021gsm8k} substantially increases error rates, while \citet{mirzadeh2024gsm} demonstrates that LLMs are far more sensitive to changes in numerical values than to changes in entity names in \texttt{GSM8K}, suggesting that LLMs often rely on pattern recall over genuine reasoning. Our observations are also conceptually related to recent work on Chain-of-Thought (CoT) faithfulness \cite{lanham2023measuring}, which shows that generated reasoning traces do not always faithfully reflect the computation leading to the final answer. \textit{In-Context Laziness}, however, reflects a distict failure mode in long-context time series reasoning: instead of executing the required computation over the full CGM record, models anchor on a rough intermediate value and apply minor narrative adjustments to justify a confident answer rather than expressing uncertainty. 

We argue that this phenomenon may stem from shortcuts acquired during post-training. In many training settings, models are primarily rewarded for producing the correct final answer rather than demonstrating complete intermediate reasoning or identifying when external tools are necessary. As a result, models may learn to generate plausible responses without fully processing the provided context.

We believe future models should be trained to better recognise their own limitations. When encountering complex tasks that exceed their reasoning or computational capabilities, models should either explicitly admit uncertainty or proactively invoke appropriate tools (or require tool use from the user) to complete the required analysis. 

\input{tables/failure_improvement}

\input{tables/failure_analysis}

\section{Data details}

\begin{figure*}[t!]
    \centering
    \begin{subfigure}[b]{0.32\linewidth}
        \centering
        \includegraphics[width=\linewidth]{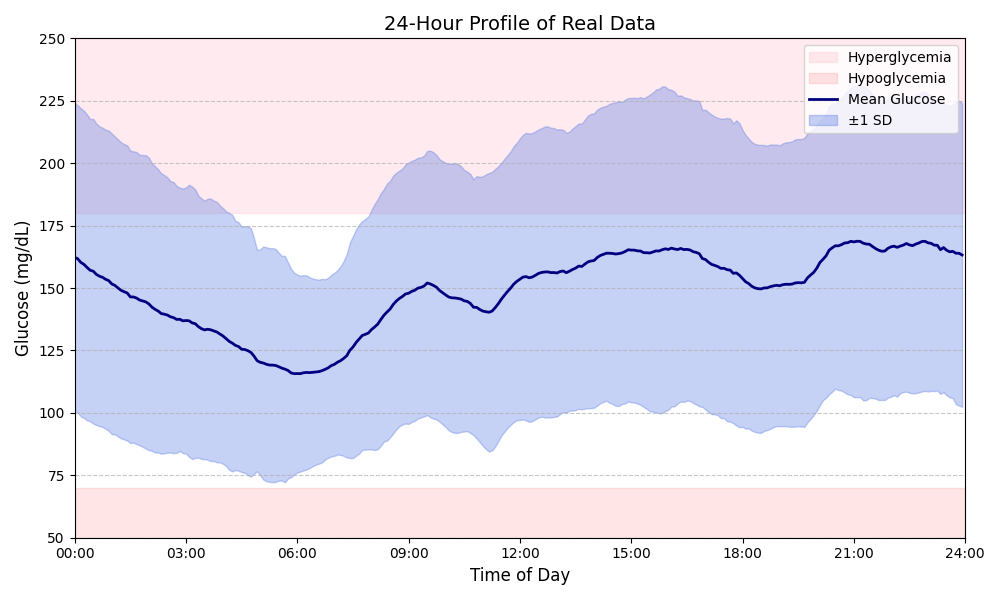}
        \caption{Real-world data}
        \label{fig:real_profile}
    \end{subfigure}%
    \hfill
    \begin{subfigure}[b]{0.32\linewidth}
        \centering
        \includegraphics[width=\linewidth]{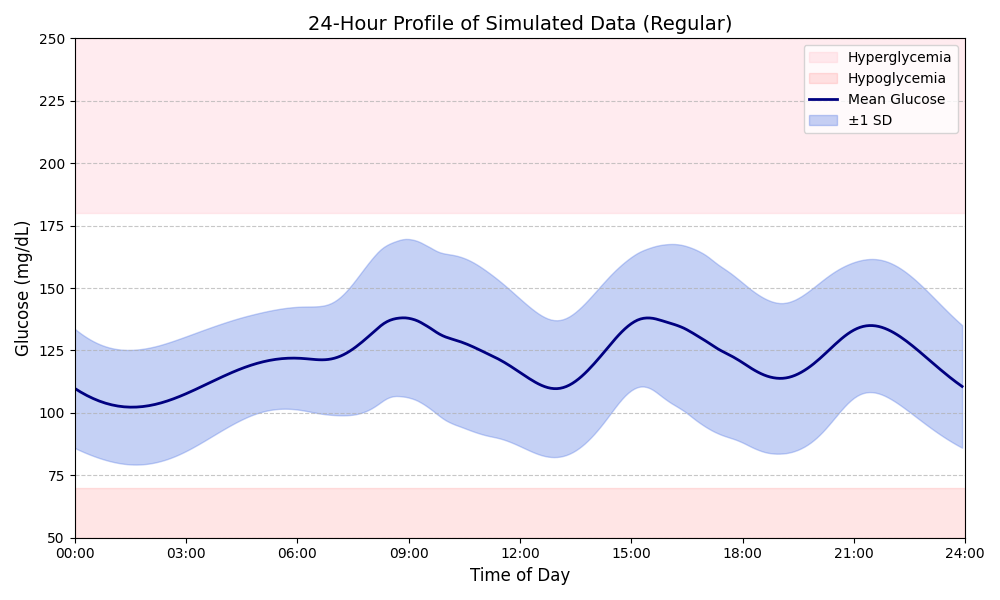}
        \caption{Simulation data (no fault injection)}
        \label{fig:sim_profile}
    \end{subfigure}%
    \hfill
    \begin{subfigure}[b]{0.32\linewidth}
        \centering
        \includegraphics[width=\linewidth]{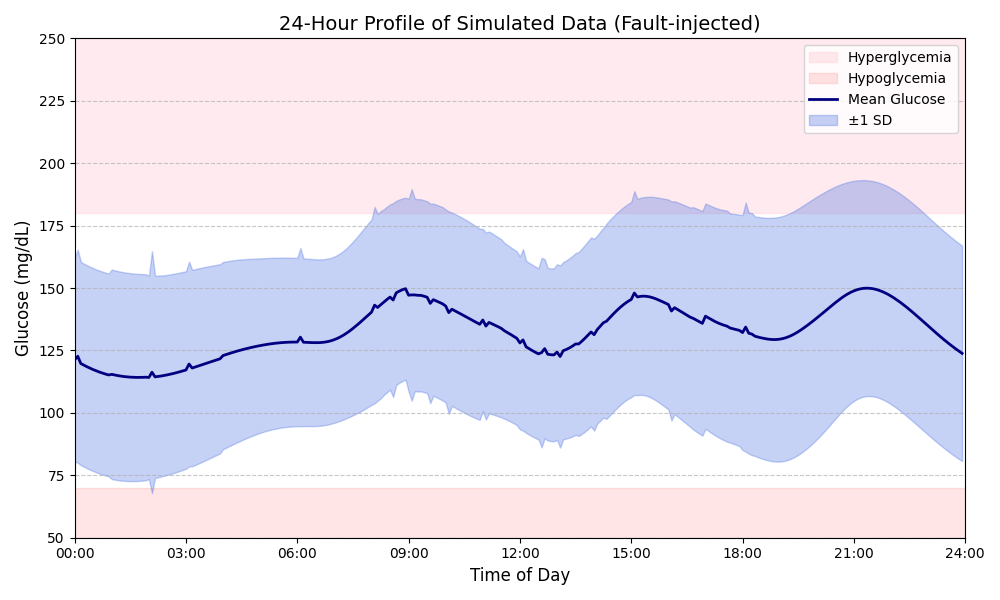}
        \caption{Fault-injected simulation data}
        \label{fig:fault_profile}
    \end{subfigure}

    \caption{Comparison of 24-hour glucose profiles across real-world, clean simulation, and fault-injected simulation data.}
    \label{fig:24hour_comparison}
\end{figure*}

\label{app:data_details}
\begin{figure}[t!]
    \centering
    \includegraphics[width=1.0\linewidth]{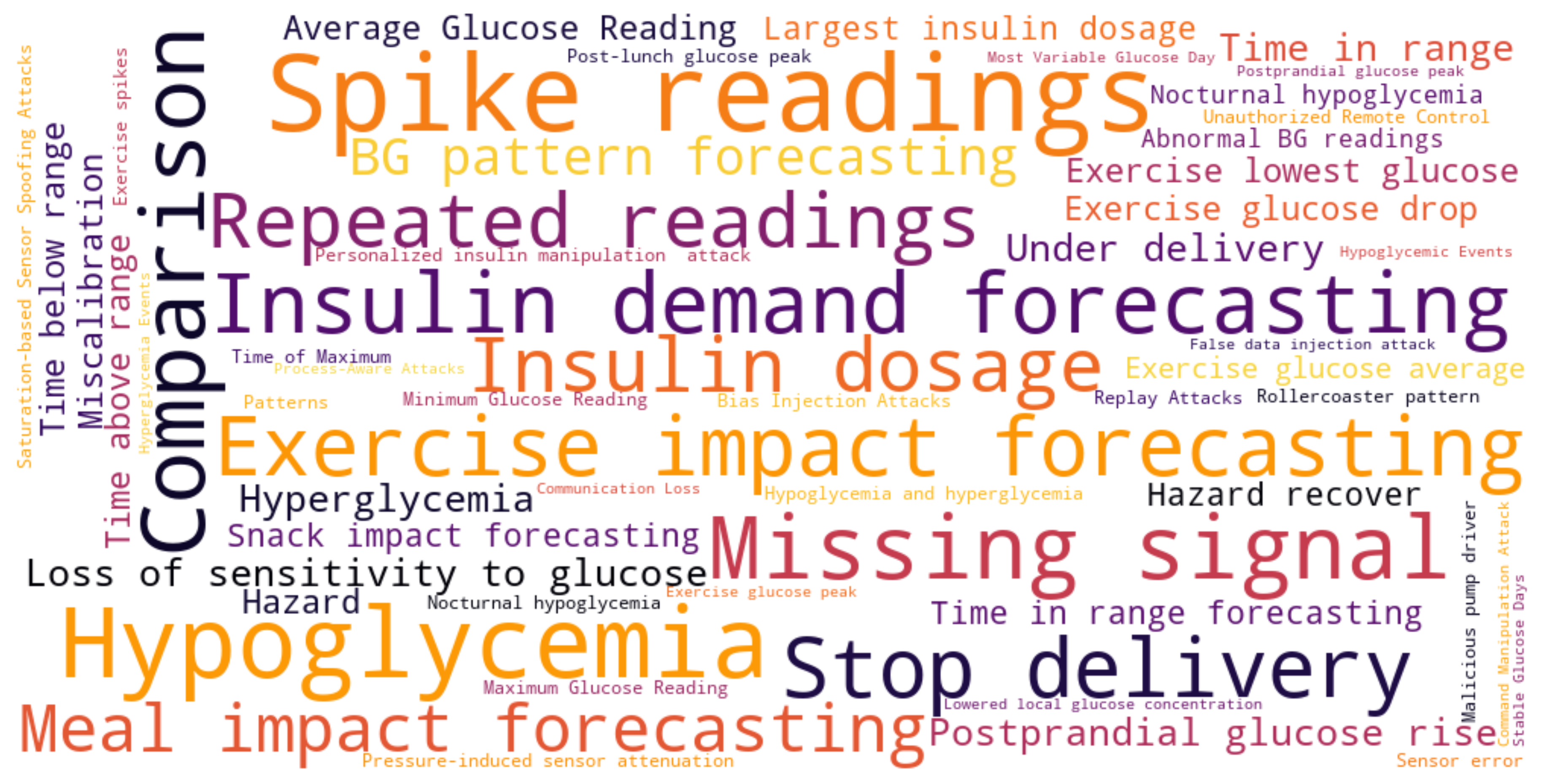}
    \caption{Semantic distribution and diversity of benchmark task prototypes.}
    \label{fig:prototype_cloud}
\end{figure}

We included 127 QA templates and summarized the data distribution in Table \ref{tab:qa_dataset_distribution}. Figure~\ref{fig:prototype_cloud} shows the most frequently used words in the word cloud image. Data simulation/collection details are explained in the following sections.

\subsection{AID System Closed-loop Simulation Testbed}
The closed-loop simulation is built on a clinically validated AID simulation testbed \cite{siket2025py}, which incorporates multiple controller strategies and a cohort of 20 virtual T1D patients equipped with the physical activity submodels \citep{rashid2019simulation}. Unlike existing simulators that often assume rigid daily routines and fault-free operation, we incorporated diverse meal and exercise arrangements and explicitly modelled 17 documented device malfunctions and cyber-physical threats \citep{kapadia2024failure,niu2025securing}, enabling evaluation under real-world operational challenges. 

We adopt the simulation testbed as the primary data source for four reasons: (1) it has been validated to capture the predominant and most frequent drivers of glycemic dynamics in response to meals, insulin delivery, and physical activity with high fidelity \citep{rashid2019simulation}; (2) it enables explicit modelling and precise labelling of faults within the control loop, allowing assessment of safety-critical reasoning related to anomaly detection and hazard mitigation without posing risks to real patients; (3) it provides a physiologically diverse virtual cohort representative of the T1D population, producing monitoring trajectories that generalise beyond isolated case studies and support evaluation across inter-individual variability; and (4) its test environment is controllable and less confounding interference factors, facilitating systematic evaluations of model strengths and weaknesses.

\subsection{Faults Modelling}
\label{appendix:faults}
To construct comprehensive and realistic fault conditions in our closed-loop simulation testbed, we synthesise fault patterns derived from documented malfunctions in glucose sensing and insulin infusion \citep{kapadia2024failure}, as well as cyber-physical attack vectors that compromise system availability and integrity \citep{niu2025securing}, all of which manifest as distortions in monitoring data. Tables~\ref{tab:malfunctions} and \ref{tab:attacks} summarise these underlying malfunction and attack categories. Building on these sources, we define 17 fault patterns and 2 physiological hazards used in our benchmark. Each pattern is implemented through explicit manipulations within the simulation loop, such as perturbing CGM values, altering basal insulin delivery, replaying historical events, or dropping sensor signals, enabling systematic evaluation of anomaly detection, fault reasoning, and safety-critical interpretability.

\input{tables/table_malfunctions}
\input{tables/table_cyberattacks}

\subsection{Data simulation settings}

It consists of three daily schedules and one with faulty injection, differing in exercise frequency and randomness, simulated over 30 days for 16 patients.

\textbf{Meal Schedule }(Both Scenarios):
\begin{itemize}[nosep]
    \item Breakfast: 45–60 g carbohydrates between 06:00–08:00
    \item AM snack: 10–20 g between 09:00–11:00
    \item Lunch: 55–60 g between 12:00–14:00
    \item PM snack: 10–20 g between 15:00–17:00
    \item Dinner: 45–60 g between 18:00–20:00
\end{itemize}
Each meal random start and lasts 30–60 minutes, while snacks last 5–15 minutes.

\textbf{Scenario 1}: Running randomly sampled between 20 and 40 minutes duration during 15:00-17:00 with speed 1.7-4.0 km/h. This dataset is used for process mining tasks. Random seed is 64. 

\textbf{Scenario 2}: Running randomly sampled between 20 and 40 minutes duration during 08:00-09:30 with speed 2.0-5.0 km/h. Cycling randomly sampled between 20 and 40 minutes duration during 16:00-18:30 with cycling power 70-90 W. A late-night snack was injected between 22:00-24:00 on day 29. An overeating lunch with carbs between 160-240 grams was injected on day 30. This dataset is used for prediction tasks. Random seed is 64. 

\textbf{Scenario 3 with faults injection}: The breakfast, lunch, and dinner carbs are randomly sampled from 50-80 g, 60-90 g, and 50-80 g. Cycling randomly sampled between 20 and 90 minutes duration during 15:00-17:00 with cycling power 60-90 W. 17 faults with a designed arrangement are injected in the simulation loop (can be found in the open-source code repository). Random seed is 146.

\subsection{Real-world and simulation data statistics}
The characteristics and statistics of real-world data, regular simulation data (without fault injection), and fault-injected simulation data are visualized in Table~\ref{tab:glucose_characteristics} and Figure~\ref{fig:24hour_comparison}. QA data schema is summarized in Table~\ref{tab:data_schema}.

\input{tables/Comparison_of_glucose_characteristics}

\input{tables/qa_data_structure}

\section{Experiment setting}
\label{app:experiment_setting}
Three models are used for the experiments: GPT-5.4, claude-sonnet-4-6 and gemini-3.1-flash-lite-preview. We only used Gemini for the prompting-based baseline due to practical reasons. We used GPT, Claude, and Gemini for the agent baselines.

\subsection{Agent Baseline}
For GPT-5.4, The agent builder framework from the OpenAI platform is used to set up the agent workflow and the code is generated from the platform which uses the OpenAI agents SDK for abstraction. The workflow has a code interpreter tool which runs in a virtual container. This container has access to the patient data in the form of a CSV file and the model is enabled to write python code to try answering the users question. A final answer extraction agent is used to extract the answer from the coding agent and save it in a structured format for evaluations. The reasoning effort is set to ``low'' and ``summary'' is enabled to extract the summary of the reasoning traces.

The official API provided by Anthropic is used for running the agent baseline for Claude-sonnet-4-6. ``Adaptive'' reasoning is enabled with ``low'' effort and maximum tokens is set to 8192. Structured output is enabled for post processing where the model will output the answer in JSON format. For Gemini flash 3.1, we use the CodeAct \cite{wang2024executable} framework to control the token consumption and setup the agent workflow. The model has access to a code interpreter, consistent with the other experimental settings. The thinking level is set to ``low'' with summaries enabled. 

\subsection{Prompt based  Baseline}
The text baseline for gemini-3.1-flash-lite-preview model is conducted through OpenRouter. The patient context like blood glucose, insulin, meal events and exercise are converted into raw text for inference. Structured output is enabled for post porcessing where the model outputs the answer in JSON format. 


\subsection{System Prompt}

In this section, we present the prompt templates used for the benchmark study.

\label{App:system-prompt}

\begin{figure*}[!htbp]
    \centering
    \begin{dialogbox}
    \tiny
{\centering \large \textbf{Prompting-based structured Template}\par}
You are a medical AI assistant analyzing diabetes management data. Based on the patient's health data, answer the following question accurately.\\

\textbf{PATIENT DATA OVERVIEW:} \\
This data represents a continuous glucose monitoring (CGM) session for a diabetes patient. The data includes:\\
 - Blood glucose readings taken every 5 minutes (normal range: 70--180 mg/dL)\\
 - Carbohydrate intake events with timing and amounts\\
 - Insulin delivery events (basal background insulin and bolus meal insulin)\\
 -  Physical activity events (running/cycling with duration and intensity)\\
 
The data may contain various artifacts, sensor issues, or abnormal patterns that need to be identified and analyzed.\\

\textbf{Patient Health Data:}
Blood Glucose Readings (mg/dL, every 5 minutes): 8640 readings \\
Values: [108.0, 108.0, ... 108.0, 129.2, 128.8] \\
\textbf{Insulin Events (90 total):} \\
 - 1 Day 1 00:00: 1.3U (basal\_insulin) \\
 ... \\
 - 5 Day 2 18:18: 3.1U (bolus\_insulin)\\
\textbf{Carbohydrate Events (150 total):} \\
 - Week 1 Day 1 07:40: 84.3g (breakfast) \\
 ... \\
 - Week 5 Day 2 18:18: 72.6g (dinner)\\
\textbf{Exercise Events (30 total):} \\
 - Week 1 Day 1 16:49: Cycling avg power 200.0 for 40 min \\
 ... \\
 - Week 5 Day 2 16:56: Cycling avg power 140.7 for 52 min\\
 
\textbf{Question:} \{question\} \\
\textbf{Instructions:} \{answer\_instruction\} \\
\textbf{Expected Answer Type:} \{answer\_type\} (e.g., float) \\
\textbf{Example Answer:} \{answer\_example\}\\
Please analyze the data carefully and provide your answer as a \textbf{json object} in the exact format specified by the answer type. Be precise and base your response only on the data provided. Conclude your analysis with: \\

{"answer": your answer here}

    \end{dialogbox}
    \caption{System prompt template for prompting-based structured baseline where glucose readings and events are listed in distinct sections.}
    \label{fig:structured-prompt}
\end{figure*}

%% file: tables/table_atomic_reasoning_abilities.tex
\begin{table*}[ht]
\centering
\footnotesize
\resizebox{0.98\linewidth}{!}{
\begin{tabularx}{\linewidth}{
    p{0.12\linewidth}  
    p{0.14\linewidth}  
    X                  
    X                  
    X                  
}
\toprule
\textbf{Analytic Stage} &
\textbf{Atomic Ability} &
\textbf{Definition} &
\textbf{Purpose} &
\textbf{Example} \\
\midrule

\multirow{2}{*}{\shortstack{Intention \\ Understanding}} 
& \textit{Question Understanding (QU)} 
& Inferring user intent, task type, and target variables from the query. 
& Provides semantic grounding for downstream reasoning. 
& Distinguishing whether the query requests forecasting, anomaly diagnosis, or description. \\

& \textit{Instruction Alignment (IA)} 
& Following task instructions, output constraints, and evaluation requirements. 
& Ensures reliable and instruction-consistent reasoning. 
& Producing outputs in specified formats or using a prescribed reasoning chain. \\
\midrule

\multirow{5}{*}{\shortstack{Cognitive \\Reasoning}} 

& \textit{Quantitative Calculation (QC)} 
& Executing numerical/statistical operations on physiological time-series data. 
& Forms accurate quantitative perception of signals. 
& Computing means, ranges, rate-of-change, or metrics (e.g., GRI). \\

& \textit{Temporal Retrieval (TR)} 
& Selecting time-specific segments of the monitoring history.  
& Focuses analysis on relevant temporal windows. 
& Retrieving data from a specific interval (e.g., “last Sunday night”). \\

& \textit{Event Retrieval (ER)} 
& Extracting data linked to events or contextual markers. 
& Grounds reasoning in event-driven physiological changes. 
& Pulling records around meals, exercise, or insulin dosing. \\

& \textit{Comparative Analysis (CA)} 
& Comparing characteristics across multiple retrieved episodes. 
& Establishes inter-episode relationships and trends. 
& Comparing QC metrics across post-meal intervals or exercise events. \\

& \textit{Pattern Recognition (PR)} 
& Identifying unknown or emergent recurrent patterns. 
& Captures higher-level regularities in physiological data. 
& Detecting nocturnal hypoglycemia or postprandial spikes. \\

& \textit{Pattern Correlation (PC)} 
& Matching known patterns or templates against observed data. 
& Checks for presence of clinically relevant patterns or fault signatures. 
& Identifying sensor drift, pump malfunction, or hypoglycemia patterns. \\
\midrule

\multirow{3}{*}{\shortstack{Task \\Acting}} 

& \textit{Process Mining (PM)} 
& Retrospective analysis of long-horizon monitoring data. 
& Supports clinical assessment and therapy evaluation. 
& Analyzing Time-in-Range and recurring hyperglycemia patterns. \\

& \textit{Anomaly Detection(AD)} 
& Detecting abnormal physiological events or device malfunctions in real time. 
& Ensures patient safety through timely alerts. 
& Detecting sensor failure, occlusion, or attack signatures. \\

& \textit{Prediction(PD)} 
& Forecasting future physiological states using historical patterns. 
& Enables proactive therapy and preventative decisions. 
& Predicting glucose trajectories to avoid hypo-/hyperglycemia. \\

\bottomrule
\end{tabularx}
}
\caption{Atomic reasoning abilities required for a Medical Wearables Monitoring Assistant.}
\label{tab:atomic_reasoning_supp}
\end{table*}

%% file: tables/table_atomatic_reasoning_path_performance.tex
\begin{table}[t]
\centering
\scriptsize
\setlength{\tabcolsep}{4pt}
\renewcommand{\arraystretch}{1.08}
\resizebox{\linewidth}{!}{%
\begin{tabular}{lrrrr}
\toprule
Ordered sequence & Gemini$_{\textit{Prompt.}}$ & Gemini$_{\textit{Agent}}$ & GPT$_{\textit{Agent}}$ & Claude$_{\textit{Agent}}$ \\
\midrule
\multicolumn{5}{l}{\textit{Simulated dataset}} \\
QC & 68.8 & 89.5 & 92.7 & \textbf{95.6} \\
CA & 34.0 & 76.6 & \textbf{95.0} & 85.0 \\
PC & 23.8 & 45.9 & \textbf{63.5} & 62.2 \\
PR & 14.9 & 40.7 & 64.4 & \textbf{64.7} \\
ER$\rightarrow$QC & 20.8 & 60.0 & \textbf{100.0} & \textbf{100.0} \\
PC$\rightarrow$QC & 57.9 & 57.3 & \textbf{64.1} & 54.0 \\
TR$\rightarrow$QC & 42.9 & 71.0 & \textbf{86.5} & 82.1 \\
QC$\rightarrow$CA & 50.0 & \textbf{75.0} & 37.5 & 50.0 \\
TR$\rightarrow$CA & 0.0 & 20.0 & \textbf{80.0} & 60.0 \\
QC$\rightarrow$TR & 58.9 & \textbf{100.0} & \textbf{100.0} & \textbf{100.0} \\
ER$\rightarrow$PC & 0.8 & 3.9 & 8.9 & \textbf{23.5} \\
TR$\rightarrow$PC & 6.9 & 47.6 & 82.8 & \textbf{83.4} \\
QC$\rightarrow$PR & 67.0 & 66.0 & \textbf{69.0} & 67.6 \\
ER$\rightarrow$TR$\rightarrow$QC & 75.3 & 94.2 & \textbf{99.0} & \textbf{99.0} \\
PR$\rightarrow$ER$\rightarrow$QC & \textbf{100.0} & 75.0 & 87.5 & 93.8 \\
TR$\rightarrow$QC$\rightarrow$CA & 39.8 & 86.5 & 84.0 & \textbf{87.7} \\
ER$\rightarrow$QC$\rightarrow$TR & \textbf{62.8} & 49.6 & 49.6 & 49.6 \\
PR$\rightarrow$QC$\rightarrow$PC & 56.6 & 74.3 & \textbf{80.8} & 77.5 \\
CA$\rightarrow$QC$\rightarrow$PR & 21.9 & 37.5 & \textbf{71.9} & 62.5 \\
ER$\rightarrow$QC$\rightarrow$PR & 60.4 & 60.0 & \textbf{64.4} & 58.2 \\
QC$\rightarrow$CA$\rightarrow$PR & 50.0 & \textbf{56.2} & \textbf{56.2} & \textbf{56.2} \\
QC$\rightarrow$ER$\rightarrow$PR & 80.3 & 76.3 & \textbf{86.9} & 51.2 \\
TR$\rightarrow$ER$\rightarrow$PR & 0.8 & 44.3 & 62.9 & \textbf{75.6} \\
ER$\rightarrow$TR$\rightarrow$QC$\rightarrow$CA & 63.0 & 79.2 & \textbf{100.0} & 96.3 \\
ER$\rightarrow$TR$\rightarrow$QC$\rightarrow$TR & 30.5 & 56.3 & \textbf{61.4} & \textbf{61.4} \\
CA$\rightarrow$QC$\rightarrow$ER$\rightarrow$PR & 37.5 & 56.2 & \textbf{87.5} & 62.5 \\
\midrule
\multicolumn{5}{l}{\textit{Real-world dataset}} \\
QC & 78.0 & 89.9 & 93.0 & \textbf{93.8} \\
PC$\rightarrow$QC & \textbf{82.8} & 58.3 & 76.6 & 68.4 \\
TR$\rightarrow$QC & 62.8 & 92.0 & \textbf{97.7} & 86.6 \\
QC$\rightarrow$CA & \textbf{100.0} & \textbf{100.0} & \textbf{100.0} & \textbf{100.0} \\
QC$\rightarrow$TR & 51.2 & \textbf{100.0} & \textbf{100.0} & \textbf{100.0} \\
QC$\rightarrow$PR & 83.6 & 86.1 & \textbf{89.1} & 86.9 \\
ER$\rightarrow$TR$\rightarrow$QC & 70.7 & 78.8 & \textbf{86.9} & 86.2 \\
TR$\rightarrow$QC$\rightarrow$CA & 32.9 & \textbf{87.7} & \textbf{87.7} & 81.1 \\
ER$\rightarrow$QC$\rightarrow$TR & 82.8 & \textbf{92.7} & 82.7 & \textbf{92.7} \\
PR$\rightarrow$QC$\rightarrow$PC & 65.5 & 86.7 & \textbf{91.3} & 86.5 \\
CA$\rightarrow$QC$\rightarrow$PR & 30.0 & 0.0 & 40.0 & \textbf{50.0} \\
ER$\rightarrow$QC$\rightarrow$PR & 50.0 & 40.0 & \textbf{80.0} & 50.0 \\
QC$\rightarrow$CA$\rightarrow$PR & \textbf{90.0} & 50.0 & 50.0 & 50.0 \\
QC$\rightarrow$ER$\rightarrow$PR & 83.0 & 69.9 & \textbf{84.2} & 79.9 \\
ER$\rightarrow$TR$\rightarrow$QC$\rightarrow$CA & 53.9 & 67.5 & \textbf{79.7} & 75.0 \\
ER$\rightarrow$TR$\rightarrow$QC$\rightarrow$TR & 56.0 & 41.4 & 53.6 & \textbf{65.2} \\
CA$\rightarrow$QC$\rightarrow$ER$\rightarrow$PR & \textbf{80.0} & 50.0 & 60.0 & 60.0 \\
\bottomrule
\end{tabular}%
}
\caption{Nested atomic reasoning performance grouped by ordered atomic reasoning paths. Entries are mean normalized scores scaled to 0--100. Best values across different model settings are bolded.}
\label{tab:nested-atomic-reasoning}
\end{table}

%% file: tables/failure_improvement.tex
\begin{table}[t!]
\centering

\resizebox{\columnwidth}{!}{
\begin{tabular}{lccc}
\toprule
\textbf{Method} & \textbf{Correct (\checkmark and \checkmark*)} & \textbf{Applicable Total} & \textbf{Accuracy} \\
\midrule
Structured 30-day   & 20 & 44 & 45.5\% \\
Structured 7-day    & 16 & 41 & 39.0\% \\
Structured 1-day    & 22 & 34 & 64.7\% \\
Timelined 30-day    & 25 & 44 & 56.8\% \\
Agent 30-day        & 34 & 44 & 77.3\% \\
\bottomrule
\end{tabular}
}
\caption{Summary of correctness across Structured (30-day, 7-day, and 1-day) baseline, Timelined
baseline (30-day), and Agent baseline (30-day) on \texttt{pm} task. \checkmark and \checkmark* counted as correct. Non-applicable cases (e.g., week level comparison) with ``--'' are excluded from the total number for 7-day and 1-day baselines.}
\label{tab:failure_improvement}
\end{table}

%% file: tables/failure_analysis.tex
\begin{table*}[hbt!]
\centering
\tiny
\begin{tabular}{c|p{2cm}|p{2cm}|p{2cm}|p{2cm}|p{2cm}|p{2.8cm}}
\toprule
\textbf{ID} & \textbf{30-day} & \textbf{7-day} & \textbf{1-day} & \textbf{Timelined} & \textbf{Agent} & \textbf{Reasoning Path} \\
\midrule

\texttt{pm\_0}  & RC; UA & RC; UA & \checkmark & RC; UA & \checkmark  & QC \\
\texttt{pm\_1}  & \checkmark & \checkmark & \checkmark & \checkmark & \checkmark  & QC \\
\texttt{pm\_2}  & \checkmark & \checkmark & \checkmark & \checkmark* & \checkmark  & QC \\
\texttt{pm\_3}  & TM; UA & TM: UA & \checkmark & \checkmark & \checkmark  & QC $\rightarrow$ TR \\
\texttt{pm\_4}  & RC; UA & \checkmark & \checkmark & \checkmark* & \checkmark  & QC \\
\texttt{pm\_5}  & RC; UA & \checkmark & \checkmark & RC; UA & \checkmark  & QC \\
\texttt{pm\_6}  & \checkmark & \checkmark & \checkmark & \checkmark & \checkmark  & QC \\
\texttt{pm\_7}  & \checkmark & TM & \checkmark & \checkmark* & \checkmark  & QC \\
\texttt{pm\_8}  & \checkmark & TM & \checkmark & \checkmark* & Failed to calculate & QC \\
\texttt{pm\_9}  & RC; UA & RC; UA & RC; UA & RC; UA & \checkmark  & QC \\
\texttt{pm\_10} & \checkmark* & \checkmark & \checkmark & \checkmark* & \checkmark  & QC \\
\texttt{pm\_11} & \checkmark & TM & \checkmark & \checkmark* & Failed to calculate & QC \\
\texttt{pm\_12} & \checkmark & RC & \checkmark & \checkmark & \checkmark & QC \\
\texttt{pm\_13} & \checkmark & \checkmark* & TM & \checkmark* & \checkmark & QC $\rightarrow$ CA \\
\texttt{pm\_14} & TM; UA & TM; UA & TM; UA & TM; UA & \checkmark & TR $\rightarrow$ QC \\
\texttt{pm\_15} & \checkmark* & TM & -- & \checkmark* & \checkmark & TR $\rightarrow$ QC $\rightarrow$ CA \\
\texttt{pm\_16} & RC; UA & RC; UA & -- & \checkmark* & \checkmark & TR $\rightarrow$ QC $\rightarrow$ CA \\
\texttt{pm\_17} & RC; UA & RC; UA & -- & RC; UA & Failed to Calculate & TR $\rightarrow$ QC $\rightarrow$ CA \\
\texttt{pm\_18} & \checkmark* & \checkmark* & UA & \checkmark* & \checkmark & TR $\rightarrow$ QC $\rightarrow$ CA \\
\texttt{pm\_19} & RC; UA & RC; UA & -- & UA; RC & UA & TR $\rightarrow$ QC $\rightarrow$ CA \\
\texttt{pm\_21} & TM; UA & TM; UA & \checkmark & TM; UA & \checkmark & ER $\rightarrow$ TR $\rightarrow$ QC \\
\texttt{pm\_22} & \checkmark* & \checkmark & \checkmark & \checkmark & \checkmark & ER $\rightarrow$ TR $\rightarrow$ QC $\rightarrow$ TR \\
\texttt{pm\_23} & \checkmark* & TM; UA & \checkmark & TM; UA & \checkmark & ER $\rightarrow$ TR $\rightarrow$ QC $\rightarrow$ CA \\
\texttt{pm\_24} & TM; UA & TM; UA & TM; UA & TM; UA & Failed to calculate & ER $\rightarrow$ TR $\rightarrow$ QC \\
\texttt{pm\_25} & UA & \checkmark & -- & \checkmark & Failed to calculate & ER $\rightarrow$ TR $\rightarrow$ QC $\rightarrow$ CA \\
\texttt{pm\_26} & TM; UA & TM; UA & \checkmark & TM; UA & \checkmark & ER $\rightarrow$ TR $\rightarrow$ QC \\
\texttt{pm\_27} & \checkmark* & TM; UA & \checkmark & TM; UA & \checkmark & ER $\rightarrow$ TR $\rightarrow$ QC $\rightarrow$ CA \\
\texttt{pm\_28} & TM; UA & TM; UA & TM; UA & \checkmark & \checkmark & ER $\rightarrow$ TR $\rightarrow$ QC \\
\texttt{pm\_29} & TM; UA & TM; UA & TM; UA & TM; UA & Failed to calculate & ER $\rightarrow$ TR $\rightarrow$ QC \\
\texttt{pm\_30} & TM; UA & TM; UA & \checkmark & TM; UA & \checkmark & ER $\rightarrow$ TR $\rightarrow$ QC \\
\texttt{pm\_31} & TM; UA & TM; UA & TM; UA & TM; UA & \checkmark & ER $\rightarrow$ TR $\rightarrow$ QC \\
\texttt{pm\_32} & TM; UA & TM; UA & TM; UA & TM; UA & \checkmark & ER $\rightarrow$ TR $\rightarrow$ QC $\rightarrow$ TR \\
\texttt{pm\_33} & TM; UA & TM; UA & TM; UA & TM; UA & Failed to calculate & ER $\rightarrow$ TR $\rightarrow$ QC \\
\texttt{pm\_34} & TM; UA & TM; UA & TM & UA & \checkmark & ER $\rightarrow$ TR $\rightarrow$ QC \\
\texttt{pm\_36} & TM; UA & TM; UA & TM; UA & TM; UA & Failed to calculate & ER $\rightarrow$ TR $\rightarrow$ QC $\rightarrow$ TR \\
\texttt{pm\_37} & UA & \checkmark & -- & \checkmark* & \checkmark & ER $\rightarrow$ TR $\rightarrow$ QC $\rightarrow$ CA \\
\texttt{pm\_38} & \checkmark & -- & -- & \checkmark & \checkmark & TR $\rightarrow$ QC $\rightarrow$ CA \\
\texttt{pm\_39} & UA & -- & -- & \checkmark* & \checkmark & TR $\rightarrow$ QC $\rightarrow$ CA \\
\texttt{pm\_40} & TM; UA & -- & -- & TM; UA & Failed to calculate & TR $\rightarrow$ QC $\rightarrow$ CA\\
\texttt{pm\_41} & \checkmark & \checkmark & \checkmark & \checkmark & \checkmark & QC \\
\texttt{pm\_42} & \checkmark & \checkmark & \checkmark & \checkmark & \checkmark & ER $\rightarrow$ QC $\rightarrow$ TR \\
\texttt{pm\_43} & \checkmark & \checkmark & \checkmark & \checkmark & \checkmark & TR $\rightarrow$ QC \\
\texttt{pm\_44} & \checkmark & \checkmark & \checkmark & \checkmark & \checkmark & TR $\rightarrow$ QC \\
\texttt{pm\_45} & \checkmark & \checkmark & -- & \checkmark & \checkmark & TR $\rightarrow$ QC $\rightarrow$ CA \\
\bottomrule
\end{tabular}
\caption{
Failure-type comparison across Structured (30-day, 7-day, and 1-day) prompts, Timelined baseline (30-day), and Agent baseline (30-day). \checkmark\ indicates correct answers; \checkmark* indicates coincidentally correct answers based on assumptions/heuristics. RC (Reluctance to Calculate); TM (Temporal Misalignment); UA (Unsupported Assumption); GU (Guessing over Uncertainty).
}
\label{tab:failure_pm_compare}
\end{table*}

%% file: tables/table_malfunctions.tex
\begin{table*}[htbp]
\centering

\resizebox{\textwidth}{!}{
\begin{tabular}{p{3.2cm} p{6.5cm} p{3.7cm} p{2.6cm} p{1.8cm}}
\toprule
\textbf{Malfunction} & \textbf{Description} & \textbf{Simulated Fault Pattern} & \textbf{Failure Mode} & \textbf{ad\_IDs} \\
\midrule
Miscalibration &
Incorrect calibration during BG fluctuations; infrequent calibration; low-value calibration during unstable diffusion. &
Positive/negative bias in BG readings &
Positive/negative biased readings &
53 \\
\addlinespace

Pressure-induced sensor attenuation &
Pressure during sleep or tight clothing causing signal drop. &
Sudden negative BG drop &
Negative spike readings &
54 \\
\addlinespace

Unphysiological motion-induced spikes &
Transient spikes exceeding physiological thresholds due to motion. &
Positive/negative spikes during activity &
Positive/negative spike readings &
55 \\
\addlinespace

Loss of sensitivity to glucose &
Improper insertion, dislocation, or signal averaging along sensor. &
(1) Zero readings; (2) Flat readings with noise; (3) Spikes &
Zero / repeated / spike readings &
8, 56 \\
\addlinespace

Lowered local glucose concentration &
Bleeding, foreign-body response, diffusion restriction. &
Average BG decreased by 32 mg/dL &
Decreased average BG level &
57 \\
\addlinespace

Communication loss &
Wireless shielding or excessive distance. &
Missing BG readings &
Missing signal &
43 \\
\addlinespace

Stop delivery &
Empty reservoir, occlusion, kinked set, or blocked cannula. &
Pump reports normal rate; model basal rate forced to 0 &
Unknown stopped delivery &
58, 59, 62, 63 \\
\addlinespace

Under delivery &
Air bubbles, foreign-body response, lipodystrophy. &
Pump reports normal rate; model basal rate reduced by 0.5 U/hr &
Unknown under delivery &
60, 61, 64 \\
\bottomrule
\end{tabular}}
\caption{Summary of CGM sensors and insulin pumps malfunctions from \citet{kapadia2024failure}.}
\label{tab:malfunctions}
\end{table*}

%% file: tables/table_cyberattacks.tex
\begin{table*}[htbp]
\centering

\resizebox{\textwidth}{!}{
\begin{tabular}{p{3.5cm} p{6.5cm} p{3.8cm} p{2.5cm} p{1.8cm}}
\toprule
\textbf{Attack Type} & \textbf{Description} & \textbf{Simulated Fault Pattern} & \textbf{Failure Mode} & \textbf{ad\_IDs} \\
\midrule
Availability attack (DoS, jamming, ransomware) &
System overload halts insulin delivery or data transmission. &
Missing CGM readings; insulin rate forced to 0 &
Missing signal &
43 \\
\addlinespace

Process-aware attack &
Altering sensory or controller logic using process knowledge. &
False meal event &
False meal &
44 \\
\addlinespace

Replay attack &
Replaying insulin requests or meal events. &
Repeated bolus; repeated BG episode &
False bolus / repeated episode &
45 \\
\addlinespace

Personalised insulin dose manipulation &
Targeted overdosing/underdosing that avoids detection. &
(1) Increased basal during exercise; (2) Bolus during activity &
Overdelivery / false bolus &
46 \\
\addlinespace

Bias injection attack &
Injecting constant bias into sensor signals. &
BG ±32 mg/dL for random interval &
Positive/negative biased readings &
47 \\
\addlinespace

Unauthorized remote control &
Unauthorized alteration of pump settings. &
Bias/max/min basal rates; false bolus &
Biased/max/min basal &
48 \\
\addlinespace

Command manipulation attack &
Tampering with controller algorithm instructions. &
Basal rate ±0.5 U/hr &
Positive/negative biased basal &
49 \\
\addlinespace

Saturation-based sensor spoofing &
Optical saturation to prevent legitimate sensing. &
BG forced to 175 mg/dL &
Maximized readings &
50 \\
\addlinespace

False data injection attack &
Replacing data with random or synthetic values. &
BG forced to 175, 80, or synthetic smooth trajectory &
Max/min/ synthesized readings &
51 \\
\addlinespace

Malicious pump driver &
Tampering with pump I/O operations. &
Basal rate forced to max or min &
Max/min basal &
52 \\
\bottomrule
\end{tabular}}
\caption{Cyber-physical attacks affecting AID systems' availability and integrity from \cite{niu2025securing}.}
\label{tab:attacks}
\end{table*}

%% file: tables/Comparison_of_glucose_characteristics.tex
\begin{table*}[t]
\footnotesize
\centering
\begin{tabular}{l ccc}
\toprule
\textbf{Metric} & \textbf{Real-World Data} & \textbf{Simulated (Regular)} & \textbf{Simulated (Fault-Injected)} \\
\midrule

Number of data points & 13,565.7 (9,843.3) & 17,280.0 (0.0) & 8,550.0 (0.0) \\
& [7,411.0, 32,294.0] & [17,280.0, 17,280.0] & [8,550.0, 8,550.0] \\
\addlinespace

Average glucose (mg/dL) & 155.5 (23.7) & 121.5 (14.0) & 132.8 (21.0) \\
& [132.2, 198.1] & [82.9, 146.1] & [115.2, 189.9] \\
\addlinespace

Glucose management indicator & 7.03 (0.57) & 6.22 (0.34) & 6.49 (0.50) \\
& [6.47, 8.05] & [5.29, 6.80] & [6.07, 7.85] \\
\addlinespace

Coefficient of variation & 0.36 (0.04) & 0.20 (0.08) & 0.24 (0.08) \\
& [0.30, 0.40] & [0.12, 0.34] & [0.16, 0.43] \\
\addlinespace

Minimum glucose (mg/dL) & 41.6 (4.4) & 58.1 (24.4) & 36.4 (8.3) \\
& [39.0, 50.5] & [18.0, 83.8] & [25.8, 48.0] \\
\addlinespace

Maximum glucose (mg/dL) & 367.1 (32.0) & 212.4 (53.0) & 282.3 (63.4) \\
& [329.7, 400.0] & [157.0, 341.9] & [206.9, 392.2] \\


Percent time sensor active & 79.0\% (24.0\%) & 100.0\% (0.0\%) & 99.0\% (0.0\%) \\
& [37.0\%, 99.1\%] & [100.0\%, 100.0\%] & [99.0\%, 99.0\%] \\
\addlinespace

Time in range (70–180 mg/dL) & 67.3\% (14.2\%) & 94.0\% (7.9\%) & 90.3\% (12.9\%) \\
& [43.1\%, 78.9\%] & [74.1\%, 100.0\%] & [55.6\%, 99.3\%] \\
\addlinespace

Time above range 1 (>180 mg/dL) & 30.2\% (15.4\%) & 2.8\% (4.6\%) & 8.5\% (12.8\%) \\
& [16.5\%, 56.2\%] & [0.0\%, 18.0\%] & [0.0\%, 43.6\%] \\
\addlinespace

Time above range 2 (>250 mg/dL) & 7.4\% (6.5\%) & 0.4\% (1.2\%) & 3.1\% (9.2\%) \\
& [2.0\%, 19.0\%] & [0.0\%, 4.6\%] & [0.0\%, 32.4\%] \\
\addlinespace

Time below range 1 (<70 mg/dL) & 2.5\% (1.7\%) & 3.2\% (6.9\%) & 1.2\% (1.5\%) \\
& [0.7\%, 4.6\%] & [0.0\%, 25.9\%] & [0.0\%, 5.4\%] \\
\addlinespace

Time below range 2 (<54 mg/dL) & 0.4\% (0.3\%) & 1.7\% (3.7\%) & 0.6\% (0.9\%) \\
& [0.1\%, 0.9\%] & [0.0\%, 13.4\%] & [0.0\%, 3.3\%] \\

\bottomrule
\end{tabular}
\caption{Comparison of glucose characteristics across real-world data, regular simulation data, and fault-injected simulation data. Values are reported as Mean (SD) [Min, Max].}
\label{tab:glucose_characteristics}
\end{table*}

%% file: tables/qa_data_structure.tex
\begin{table*}[htbp]
\centering
\begin{small}

\begin{tabular}{@{} p{2.4cm} p{2.8cm} p{5.2cm} >{\ttfamily\footnotesize\raggedright\arraybackslash}p{4.2cm} @{} }

\toprule
\textbf{Key / Field} &
\textbf{Sub-Key} &
\textbf{Data Type \& Internal Structure} &
\textbf{Concrete Runtime Example} \\
\midrule

patient\_id & --- & string & "Patient\_1" \\

\midrule
\rowcolor[gray]{0.95}
\multicolumn{4}{@{}l}{\textbf{input\_context} --- Longitudinal Physiological, Therapeutic, and Behavioral Context Logs} \\
\midrule

input\_context & bg\_mgdl &
list[float] (CGM readings every 5 min in mg/dL) &
[110.7, 112.1, 113.9] \\
\cmidrule{2-4}

& carb\_events &
list[dict] \{time: float (min),  day: int,  time\_str: string, carbs: float (g), meal\_type: string\} &
[\{"time": 363.64, "day": 1, "time\_str": "06:03", "carbs": 74.2, "meal\_type": "breakfast"\}] \\
\cmidrule{2-4}

& cycling\_events &
list[dict] \{time: float (min),  day: int,  time\_str: string, duration: float (min), magnitude: float (W), exercise\_type: string\} &
[\{"time": 16871.52, "day": 12, "time\_str": "17:11", "duration": 37.37, "magnitude": 78.9, "exercise\_type": "cycling"\}] \\
\cmidrule{2-4}

& insulin\_events &
list (mU/min) or list[dict] \{time: float (min),  day: int,  time\_str: string, dosage: float (mU/min), insulin\_type: string\}
 &
[55.1, 57.6, 60.7] or [\{"time": 25415, "day": 18, "time\_str": "15:35", "dosage": 44.2, "insulin\_type": "basal\_insulin"\}] \\
\cmidrule{2-4}

& running\_events &
list[dict] \{time: float (min),  day: int,  time\_str: string, duration: float (min), magnitude: float (km/h), exercise\_type: string\} &
[\{"time": 32225.45, "day": 23, "time\_str": "09:05", "duration": 31.11, "magnitude": 2.3, "exercise\_type": "running"\}] \\

\midrule
\rowcolor[gray]{0.95}
\multicolumn{4}{@{}l}{\textbf{qa\_pairs} --- Evaluation Benchmark Target Fields} \\
\midrule

qa\_pairs & question\_id &
Unique serialization identifier &
"pd\_1" \\
\cmidrule{2-4}

& question\_text &
Natural language query &
"Predict what my blood glucose level will be in 30 minutes?" \\
\cmidrule{2-4}

& question\_prototype &
Abstract semantic task type &
"Short-term BG forecasting" \\
\cmidrule{2-4}

& answer &
Ground-truth &
66.7 \\
\cmidrule{2-4}

& answer\_type &
Target answer format &
"float" \\
\cmidrule{2-4}

& example\_answer &
Randomly generated answer for format reference &
39.35 \\
\cmidrule{2-4}

& cognitive\_level &
Cognitive level of descriptive, memory, or pattern &
"Pattern" \\
\cmidrule{2-4}

& cognitive\_atomic &
Involved atomic reasoning capabilities &
"PR,QC,PC" \\
\cmidrule{2-4}

& function\_name &
Internal programmatic ground-truth generation function &
"get\_future\_bg\_values" \\
\cmidrule{2-4}

& answer\_generation\_rule &
Logic of answer extraction &
"The blood glucose value 30 minutes after the cutting point." \\
\cmidrule{2-4}

& answer\_instruction &
Further explain the question and procedures to derive the answer &
"Use the relevant information given and predict a valid blood glucose value" \\
\cmidrule{2-4}

& metric &
Metric used in the evaluation &
"MAE" \\
\cmidrule{2-4}

& patient\_id &
Patient cross-reference index &
"Patient\_1" \\

\bottomrule
\end{tabular}

\end{small}
\caption{QA data schema and specification.}
\label{tab:data_schema}
\end{table*}